%% file: cas-dc-template.tex
\documentclass[a4paper,fleqn]{cas-dc}

\usepackage[authoryear,longnamesfirst]{natbib}

\usepackage{amssymb,mathtools,siunitx,booktabs,multirow,footmisc}

\newcommand{\V}{\mathcal{V}}
\newcommand{\F}{\mathcal{F}}
\newcommand{\R}{\mathbb{R}}
\newcommand{\X}{\mathcal{X}}
\newcommand{\B}{\mathrm{B}}
\newtheorem{definition}{Definition}
\newcommand{\E}{\mathcal{E}}
\renewcommand{\quote}[1]{“#1”}
\newcommand{\I}{\mathrm{I}}
\newcommand{\SO}{\mathrm{SO}}
\newcommand{\T}{\mathsf{T}}
\newcommand{\U}{\mathcal{U}}
\newcommand{\N}{\mathcal{N}}
\newcommand{\J}{\mathrm{J}}
\renewcommand{\L}{\mathrm{L}}
\newcommand{\M}{\mathcal{M}}
\newcommand{\SE}{\mathrm{SE}}
\newtheorem{proposition}{Proposition}
\newcommand{\res}[2]{\left.#1\right|_{#2}}
\newtheorem{remark}{Remark}
\newtheorem{corollary}{Corollary}

\renewcommand{\cite}{\citep}

\newcommand{\revision}[1]{\textcolor{black}{#1}}
\usepackage[normalem]{ulem}
\newcommand{\revisionout}[1]{}
\newcommand{\revvision}[1]{\textcolor{black}{#1}}
\newcommand{\revvvision}[1]{\textcolor{black}{#1}}
\newcommand{\revvvisionout}[1]{}

\def\tsc#1{\csdef{#1}{\textsc{\lowercase{#1}}\xspace}}
\tsc{WGM}
\tsc{QE}

\newproof{proof}{Proof}

\begin{document}
\let\WriteBookmarks\relax
\def\floatpagepagefraction{1}
\def\textpagefraction{.001}

\shorttitle{SE(3)-Equivariant Hemodynamics Estimation}    

\shortauthors{Suk et al.}  

\title [mode = title]{Mesh Neural Networks for SE(3)-Equivariant Hemodynamics Estimation on the Artery Wall}  



%

\author[1]{Julian Suk}[orcid=0000-0003-0729-047X]

\cormark[1]


\ead{j.m.suk@utwente.nl}



\affiliation[1]{organization={Department of Applied Mathematics \& Technical Medical Center, University of Twente},
            city={Enschede},
            postcode={7522 NB}, 
            country={The Netherlands}}

\author[2,3]{Pim de Haan}[]





\affiliation[2]{organization={Qualcomm AI Research, Qualcomm Technologies Netherlands B.V.},
            city={Nijmegen},
            postcode={6546 AS}, 
            country={The Netherlands}}

\author[3]{Phillip Lippe}[]





\affiliation[3]{organization={QUVA Lab, University of Amsterdam},
            city={Amsterdam},
            postcode={1012 WX}, 
            country={The Netherlands}}

\author[1]{Christoph Brune}[]

\author[1]{Jelmer M. Wolterink}[]

\cortext[1]{Corresponding author}



\begin{abstract}
Computational fluid dynamics (CFD) is a valuable asset for patient-specific cardiovascular-disease diagnosis and prognosis, but its high computational demands hamper its adoption in practice. Machine-learning methods that estimate blood flow in individual patients could accelerate or replace CFD simulation to overcome these limitations. In this work, we consider the estimation of vector-valued quantities on the wall of three-dimensional geometric artery models. We employ group-equivariant graph convolution in an end-to-end
SE(3)-equivariant
neural network that operates directly on triangular surface meshes and makes efficient use of training data. We run experiments on a large dataset of synthetic coronary arteries and find that our method estimates directional wall shear stress (WSS) with an approximation error of 7.6\% and normalised mean absolute error (NMAE) of 0.4\% while up to two orders of magnitude faster than CFD. Furthermore, we show that our method is powerful enough to accurately predict transient, vector-valued WSS over the cardiac cycle while conditioned on a range of different inflow boundary conditions. These results demonstrate the potential of our proposed method as a plugin replacement for CFD in the personalised prediction of hemodynamic vector and scalar fields.
\end{abstract}




\maketitle

\input{content}

\appendix

\input{appendix}

\section*{Acknowledgements}

This work is funded in part by the 4TU Precision Medicine programme supported by High Tech for a Sustainable Future, a framework commissioned by the four Universities of Technology of the Netherlands. Jelmer M. Wolterink was supported by the NWO domain Applied and Engineering Sciences VENI grant (18192). This work made use of the Dutch national e-infrastructure with the support of the SURF Cooperative using grant no. EINF-2675.


\bibliographystyle{cas-model2-names}

\bibliography{references}

%

\end{document}

%% file: content.tex
\input{content/introduction}

\input{content/datasets}

\input{content/method}

\input{content/experiments}

\input{content/conclusion}

%% file: content/introduction.tex
\section{Introduction}\label{sec:introduction}

Computational fluid dynamics (CFD) is ubiquitous in science and engineering. In medicine, it allows for patient-specific, \textit{non-invasive} estimation of functional quantities related to blood flow \revvision{in 3D models that are manually or automatically extracted} from static cardiac computed tomography (CT)~\cite{TaylorFonte2013} or magnetic resonance imaging (MRI)~\cite{Steinman2002}. Hemodynamic scalar or vector fields (e.g. pressure or velocity) computed by CFD are valuable biomarkers for diagnosis~\revvision{\cite{DriessenDanad2019,TaylorPetersen2023}},
prognosis~\revvvision{\cite{BarralElSanharawi2021,CandrevaPagnoni2022}},
or treatment planning in patients with cardiovascular disease~\cite{ChungCebral2015} \revvision{and have strong potential}.
CFD simulation can be used to compute localised quantities on the artery wall, such as wall shear stress (WSS), i.e. the force exerted by the blood flow on the artery wall in tangential direction. WSS is a highly localised physical quantity that has been shown to correlate with local atherosclerotic plaque development and arterial remodelling in patients suffering from atherosclerosis~\cite{SamadyEshtehardi2011}.
Patient-specific local WSS values could be used to assess atherosclerosis risk in healthy, diseased, and stented arteries~\cite{GijsenKatagiri2019}.

While CFD has a strong potential as an \textit{in-silico} replacement for \textit{in-vivo} measurement of hemodynamic fields~\revvvision{\cite{PeperSchaap2021}}, it also has some practical drawbacks. High-quality CFD simulations require fine discretisation of the spatial and temporal domains, leading to long computation times~\cite{TaylorFonte2013}.
The time-intensive nature of high-fidelity CFD simulations limits their applicability in, e.g., virtual surgery planning or shape optimisation of medical devices~\cite{Marsden2013}. There is a practical need for fast but accurate estimation of hemodynamics.
Recent works have shown that there is great potential in deep neural networks \revvision{in} cardiovascular biomechanics modelling~\cite{ArzaniWang2022}. \revvision{One application of neural networks in hemodynamics modelling is the use of physics-informed neural networks (PINNs), in which a neural network is optimised to represent the desired hemodynamic field of a patient under physical constraints~\cite{ArzaniWang2021,RaissiYazdani2020}. However, PINNs and their graph-based variants~\cite{GaoZahr2022,ShuklaXu2022} do not naturally generalise to other patients and require per-instance optimisation. This can be as time-consuming as CFD, where multiple systems of equations have to be solved \textit{online} for each new artery. In contrast, we follow the approach of fast, generalising surrogate models. The core idea behind these models is that the time-consuming computation is moved \textit{offline} while hemodynamics estimation \textit{online} is fast.}
Training data can be generated using high-accuracy CFD simulations and then used to optimize a neural network that, once trained, estimates hemodynamics in a new artery in a single forward pass through the network, leading to significant speed-up.



Machine learning methods for hemodynamic parameter estimation can be subdivided into three categories. A first category is formed by \textit{parameterisation and projection} methods that re-parameterise or project the 2D artery-wall manifold from 3D to a Cartesian 1D or 2D domain and operate on this domain. This category includes approaches which use \revision{multilayer perceptrons (MLP)} to estimate (scalar) \revision{fractional flow reserve (FFR)} along the artery centerline based on shape descriptors~\cite{ItuRapaka2016}, use \revision{convolutional neural networks (CNN)} to estimate (scalar) WSS magnitude based on uniform shape sampling~\cite{SuZhang2020}, use CNNs to estimate (scalar) time-averaged and transient WSS magnitude based on a cylindrical parametrisation of the vessel wall~\cite{GharleghiSamarasinghe2020,GharleghiSowmya2022}, or use CNNs to estimate vector-valued WSS based on a cylindrical parameterisation plus uniformly sampled projections of the velocity field at several distances from the artery wall of the aorta reconstructed from 4D flow MRI~\cite{FerdianDubowitz2022}. Parameterisation and projection methods have the disadvantage that they cannot necessarily be adapted to more complex artery shapes and might fail in cases with severe pathology (e.g. aneurysms).

A second category is formed by 3D \textit{point-cloud} methods that use MLPs on points representing the native geometry of the artery. Point-cloud methods have been widely used for classification, detection, and segmentation tasks~\cite{GuoWang2021,QiYi2017}. In hemodynamic field estimation, they have been used to estimate pressure and vector-valued velocity fields on 3D point clouds~\cite{LiangMao2020}, and estimate vector-valued hemodynamic fields~\cite{LiWang2021}. Even though point-cloud methods excel at learning spatial relations from geometric data, they disregard an important part of information that is available in surface representations of arteries: the surface connectivity and curvature.

A third category of approaches exploits \textit{mesh-based} methods that use graph convolutional network (GCN) architectures and incorporate information on artery-wall structure. Mesh-based approaches incorporate additional local geometry information from the mesh in addition to the point coordinates. For example, Morales Ferez et al.~\cite{MoralesFerezMill2021} used the surface normal vector and connectivity to construct input features to a GCN predicting (scalar) endothelial cell activation potential on the left atrial appendage surface. A shortcoming of this approach is that the network predictions depend on the embedding of the mesh vertex normals in 3D Euclidean space, but the quantity of interest only depends on the intrinsic shape of the mesh. Thus, predictions are sensitive to orientation of the input and shape alignment is required.

\input{figures/zero}

In this work, we propose a mesh-based approach that processes signals intrinsically on the artery wall (Fig.~\ref{fig:zero}) \revvision{while handling meshes with variable numbers of vertices and connectivity}. The proposed method is informed by mesh properties and does not depend on the embedding of local geometry descriptors in 3D. Instead, it is invariant to translations and equivariant to rotations of the mesh. This means that vector-valued quantities like WSS rotate with the artery wall. This is data-efficient, as a single training sample covers all possible rotations and shifts of that artery and no data augmentation is required during training. Furthermore, our method is informed by anisotropic spatial interactions on the mesh, giving our filters high expressive capacity.

A preliminary version of this method was presented in \cite{SukHaan2022}, where we estimated steady-flow WSS with fixed boundary conditions. However, temporally multi-directional WSS acts as clinical biomarker for coronary plaque development~\cite{HoogendoornKok2019} and different patients have distinct coronary blood flow which
influences the WSS.
Here, we substantially extend our method to also estimate pulsatile-flow WSS and to adapt its estimation based on a given boundary condition. We present results indicating that our GCN can perform some mild \textit{extrapolation} beyond boundary conditions contained in the training data. Furthermore, we formally prove the empirical result that our method is end-to-end equivariant under rotation and translation, provide thorough experimental analysis on the influence of receptive field and sensitivity to remeshing, and include additional baseline experiments.

%% file: figures/zero.tex
\begin{figure}
    \centering
    \includegraphics[width=\columnwidth]{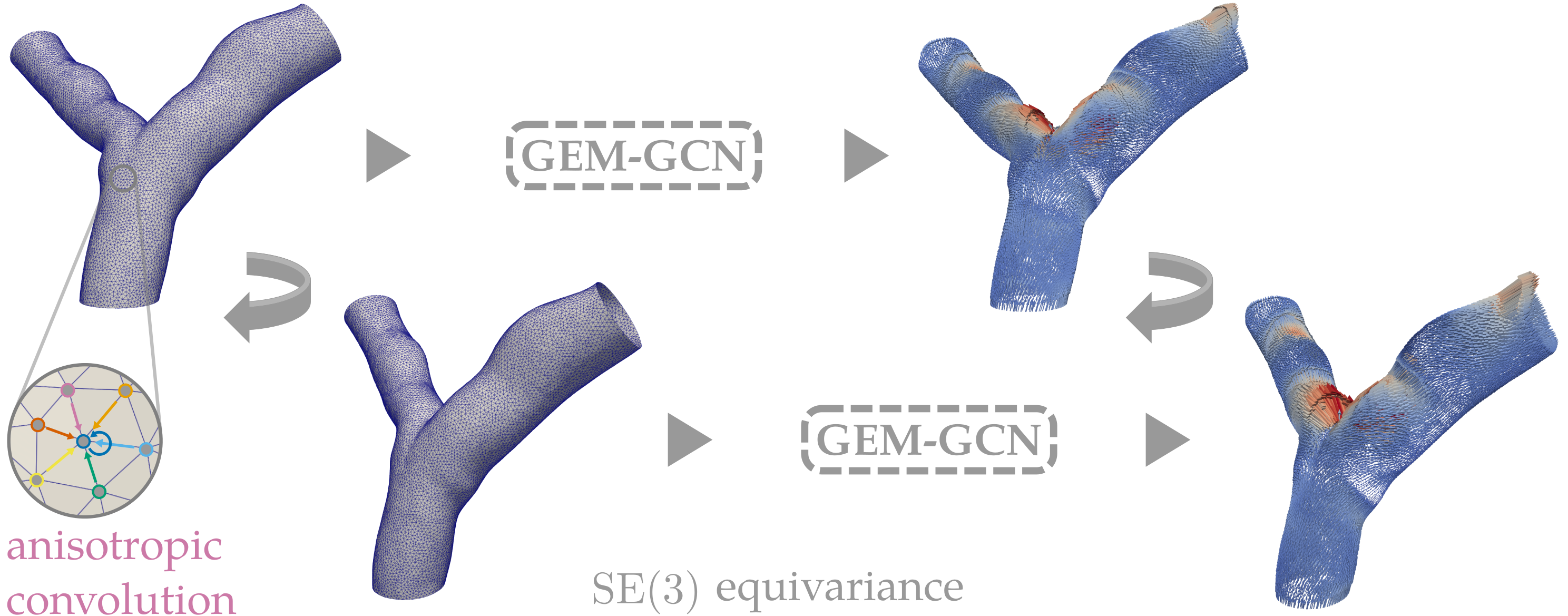}
    \caption{\textbf{Overview.} We propose a gauge-equivariant mesh-graph convolutional network (GEM-GCN) to estimate discrete hemodynamic fields mapped to the vertices of a surface mesh of the artery wall. The GCN is powered by anisotropic (spatially-oriented) gauge-equivariant mesh (GEM) convolution with high filter expressivity. The combination of GEM convolution with appropriate input features leads to an end-to-end $\SE(3)$-equivariant neural network.}
    \label{fig:zero}
\end{figure}

%% file: content/datasets.tex
\section{Data}\label{sec:data}

We propose a general method for hemodynamic field estimation on artery walls and demonstrate its value in coronary arteries, which are a key application domain for CFD. 

\subsection{Artery geometry synthesis}
We synthesise two distinct classes of representative 3D models with different topology (Fig.~\ref{fig:datasets}) for training and validation of our GCN. The first class consists of idealised, single-outlet arteries with stenoses at random locations. The second class consists of bifurcating arteries and is used to demonstrate the versatility of our method for more complex geometries as may be encountered in real-life.

\subsubsection{Single arteries}\label{subsec:single}

Emulating the shapes used in \cite{SuZhang2020}, we generate synthetic coronary arteries with a single inlet and a single outlet (Fig.~\ref{fig:datasets}). The artery centerline is defined by control points spaced at fixed increments along the horizontal axis and random uniform increments along the vertical axis in a fixed 2D plane embedded in 3D. The resulting 3D models are symmetric to that plane. We assume that the lumen contour is circular and sample its base radius $r$ from a uniform distribution $r \sim \U(1.25, 2.0)$ mm, roughly corresponding to \cite{SuZhang2020}. We randomly introduce up to two stenoses which consist of a randomly determined narrowing of up to 50~\% of the diameter, asymmetrically distributed between the top and the bottom vessel wall. The generated lumen contours are then lofted to create a watertight polygon mesh. The mesh is refined proportionally to the vessel radius along the artery centerline
to give flow-critical regions finer spatial resolution for fluid simulation. Analogously to \cite{SuZhang2020}, we add flow extensions to the inlet and outlet, whose length is five times the vessel diameter. \revision{These flow extensions are only used during simulation and later removed when simulation data is used to train and validate the deep learning model.} The shape synthesis is implemented using SimVascular~\cite{LanUpdegrove2018}.

\subsubsection{Bifurcating arteries}\label{subsec:bifurcating}

We construct the bifurcating artery models using an atlas of coronary shape statistics~\cite{MedranoGraciaOrmiston2016,MedranoGraciaOrmiston2017}. In the left main coronary bifurcation, the proximal main vessel (PMV) splits up into distal main vessel (DMV) and side branch (SB). The bifurcation can be fully described by the angles $\beta$ between centerlines of the branches DMV and SB and $\beta'$ between the bisecting line of the bifurcation and the centerline of SB (Fig.~\ref{fig:datasets}). We sample angles and lumen diameters from the atlas and use them to construct lumen contours.
\revision{Appendix~\ref{app:bifurcating} provides a detailed overview of this process. In particular, these shapes are not symmetric to any plane and cross-sections are elliptical.}
Subsequently, the generated lumen contours are lofted to create a solid polygon model, merged, and meshed. After blending of the bifurcation region to produce a more natural transition, the final surface mesh is created in a refining meshing step. The entire shape synthesis is implemented with the SimVascular Python shell.

\input{figures/datasets}

\subsection{Blood-flow simulation}

For each \revision{triangular} surface mesh, a tetrahedral volume mesh is created with five tetrahedral boundary layers (Fig.~\ref{fig:datasets}). We simulate steady and pulsatile blood flow in these meshes using the SimVascular solver for the three-dimensional, incompressible Navier-Stokes equations
\begin{align*}
    \varrho \left( \frac{\partial u}{\partial t} + (u \cdot \nabla) u \right) &= -\nabla p + \mu \Delta u \\
    \nabla \cdot u &= 0
\end{align*}
where $u \colon \Omega \to \R^3$ is the fluid velocity and $p \colon \Omega \to \R$ is the pressure in the spatial domain $\Omega$ of the artery. Dynamic viscosity and blood density are assumed to be $\mu = 0.04$ $\frac{\text{g}}{\text{cm} \cdot \text{s}}$ and $\varrho = 1.06$ $\frac{\text{g}}{\text{cm}^3}$, respectively. We model the blood vessel as rigid and apply a no-slip boundary condition, i.e. the velocity is zero at the lumen wall at all times. The inlet velocity profile is uniform for the \revision{single} arteries \revision{with the flow extensions enabling development of more realistic flow in the relevant region. To accelerate CFD simulations in the more complex and detailed bifurcating arteries, we omit flow extensions and use a parabolic profile. The inlet velocity} follows a pulsatile waveform,
scaled so that the coronary blood flow agrees with measurements in female and male patients (myocardial perfusion~\cite{PatelBui2016} times myocardial mass~\cite{CorradiMaestri2004}). A constant heart rate of 80 $\frac{1}{s}$ is used across all simulations. We model the artery outlets of the bifurcating arteries as an RCR (\quote{Windkessel}) system consisting of proximal and distal resistances and intermediate capacitance. The total applied resistance and capacitance is tuned to agree with realistic values for pressure.
The simplified boundary conditions for the steady simulations are $u_\text{in} = 20$ $\frac{\text{cm}}{\text{s}}$ for the \revision{single} and $u_\text{in} = 11.8$ $\frac{\text{cm}}{\text{s}}$ \revision{(average over the inlet)} for the bifurcating arteries as well as a pressure of $p_\text{out} = 100$ mmHg $\approx$ 13.332 kPa weakly applied at the outlet by an outlet resistance.

\revision{The Reynolds number differs per geometry and boundary condition, but we compute an estimate which we define as $\mathrm{Re} = \frac{\rho u_\mathrm{mean} 2r_\mathrm{max}}{\mu}$ with mean velocity $u_\mathrm{mean}$ across steady and pulsatile flow and maximum vessel radius $r_\mathrm{max}$ for both the single and bifurcating artery class. For the single arteries $\mathrm{Re} \approx 70$ and for the bifurcating arteries $\mathrm{Re} \approx 90$, suggesting laminar flow in both cases.}
The WSS, which we denote as $\tau$, is defined as the force exerted on the lumen wall $\partial\Omega$ by the blood flow in tangential direction and can be computed from the resulting velocity field near the lumen wall. It linearly depends on fluid velocity $u$, assuming blood to be a Newtonian fluid:
\begin{equation*}
	\tau \colon
	\begin{cases}
	\partial\Omega \to T\partial\Omega \\
	x \mapsto \mu \res{\J_u(x) \vec{n}(x)}{\perp \vec{n}}
	\end{cases}
\end{equation*}
where $T\partial\Omega$ denotes the tangent bundle of $\partial\Omega$, $\J_u$ the Jacobian of $u$, $\vec{n} \colon \partial\Omega \to \R^3$ the unit surface normal on the lumen wall and $\res{\cdot}{\perp \vec{n}}$ the perpendicular projection to $\vec{n}$.

The single-artery surface meshes have around 8,000 vertices and 17,000 triangular faces and the bifurcating artery surfaces meshes have around 17,000 vertices and 32,000 triangular faces. For an individual artery, steady-flow simulations take 10 to 24 min on an Intel Xeon Gold 5218 (16 cores, 22 MB cache, 2.3 GHz) and pulsatile-flow simulations take up to 1.6 h parallelised over 128 threads on a high-performance computing cluster. The resulting steady-flow datasets contain simulations for 2000 single arteries as well as 2000 bifurcating arteries. In addition, we \revision{create} a dataset of pulsatile-flow simulations in 731 \revision{new single arteries which are generate independently of the steady-flow case.} Note that the boundary conditions are fixed across samples and thus inherently encoded in these datasets. Therefore, we also generate pulsatile-flow datasets with varying boundary conditions, containing 187 and 117 \revision{unique} geometric models for single and bifurcating arteries, respectively. In this set, simulations for each artery are run with five random-uniform coronary blood flow values from the interval $[1.87, 4.36]$ $\frac{\text{ml}}{\text{s}}$. \revision{These values represent the average flow rate over the cardiac cycle}\revvision{. Like \citet{BeierOrmiston2016}, we work with a standard inflow profile. We} \revision{obtain the pulsatile waveform for each value by multiplying the template waveform (Fig.~\ref{fig:waveform}) with a linear factor. Determining this linear factor requires solving a nonlinear equation.} We run additional simulations with two values from $[0.63, 1.87]$ $\frac{\text{ml}}{\text{s}}$ and $[4.36, 5.61]$ $\frac{\text{ml}}{\text{s}}$, respectively, for 19 single arteries. In total, our simulation data encompasses 5,035 CFD simulations with a total runtime of ca. 2800 h.

\input{figures/waveform}

%% file: figures/datasets.tex
\begin{figure}
    \centering
    \includegraphics[width=\columnwidth]{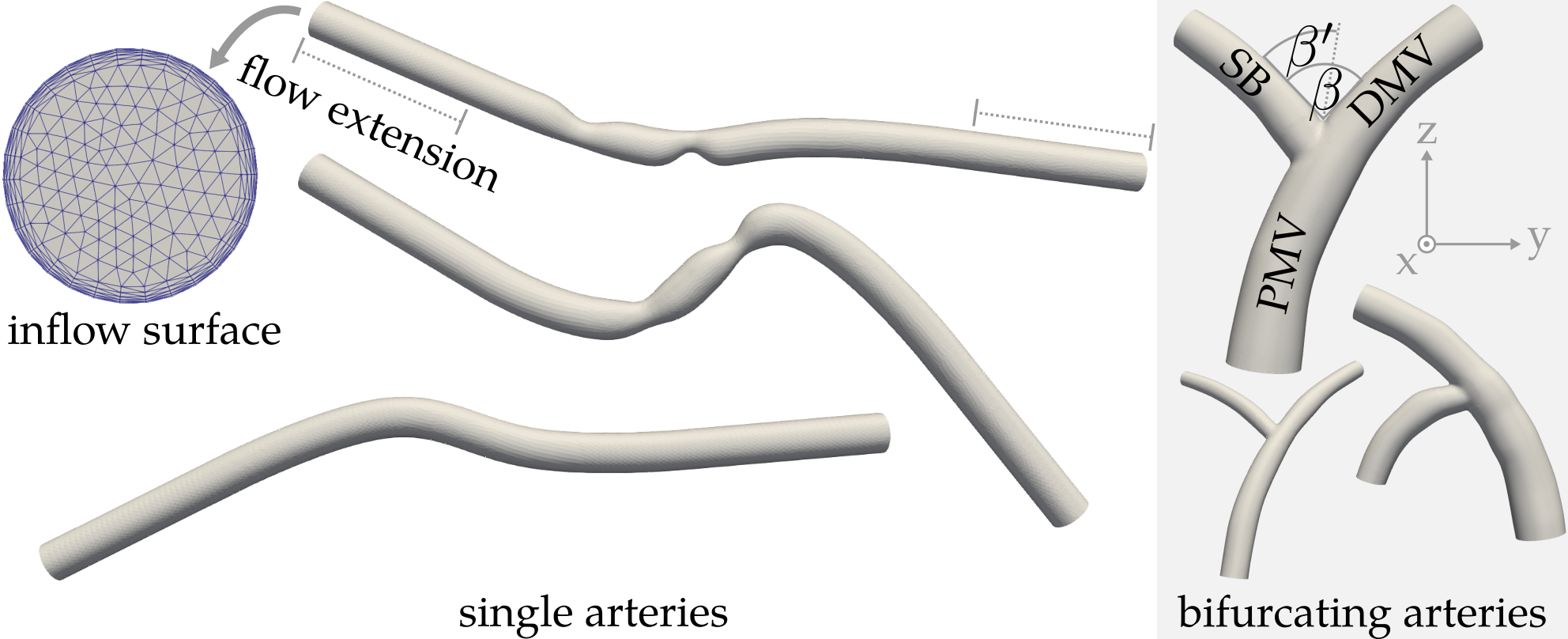}
    \caption{\textbf{Artery datasets.} We develop and evaluate our method using two distinct classes of geometric models: synthetic single arteries (left) and bifurcating arteries modelled after the left main bifurcation of the coronary artery tree (right). The single arteries contain flow extensions to let the flow fully develop from a uniform inflow boundary condition. The bifurcating arteries are simulated with parabolic inflow and thus without flow extensions. They consist of the proximal main vessel (PMV) that branches into distal main vessel (DMV) and side branch (SB). Each bifurcation can be described by the angles $\beta$ and $\beta'$.}
    \label{fig:datasets}
\end{figure}

%% file: figures/waveform.tex
\begin{figure}
    \centering
    \includegraphics[width=\columnwidth]{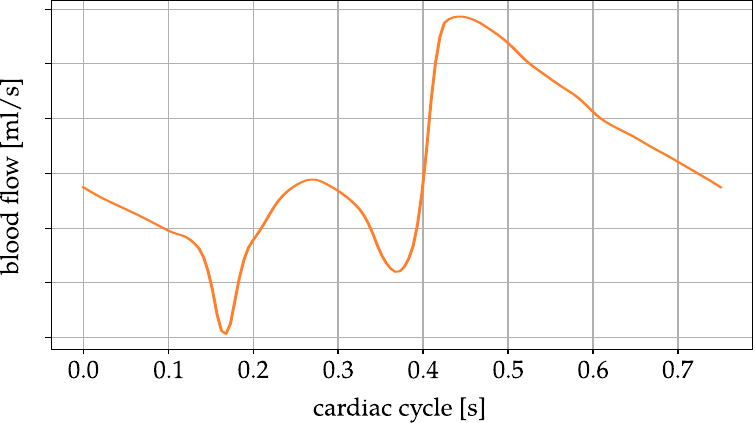}
    \caption{\revision{\textbf{Pulsatile-flow waveform}} adapted from \cite{BeierOrmiston2016}. We linearly scale this waveform for the simulations with varying (average) coronary blood flow boundary condition.}
    \label{fig:waveform}
\end{figure}

%% file: content/method.tex
\section{Learning on 3D surface meshes}\label{sec:method}

\input{figures/overview}

We propose a neural network that can estimate hemodynamic fields in the data described in Sec.~\ref{sec:data}. At the core of our approach is the assumption that hemodynamics, in the laminar regime, depend in good approximation on local artery-wall curvature, flow direction, and flow boundary conditions. As is common in CFD (see also Sec.~\ref{sec:data}), we represent the artery wall as a triangular surface mesh. Let $\Omega \subset \R^3$ be the arterial lumen and $\partial\Omega$ its 2-dimensional boundary, the artery wall. The surface mesh $\M$ is a discretisation of $\partial\Omega$ that can be fully described by a tuple of vertices and faces $\M = (\V, \F)$. We use the same mesh $\M$ from the CFD simulation to construct input features to a GCN which in turn outputs a scalar or vector for each vertex in the mesh, making use of local spatial interactions on the mesh $\M$ (Fig.~\ref{fig:zero}).

\subsection{Network architecture}\label{sec:architecture}

We propose a mesh-based GCN that takes as input a scalar or vector field of features mapped to the vertices $f^\text{in} \colon \V \to \R^{c_\text{in}}$ and outputs scalar or vector-valued predictions $f^\text{out} \colon \V \to \R^{c_\text{out}}$ mapped to the same vertices. Fig.~\ref{fig:overview} visualises the network architecture used in our experiments. The GCN is composed of convolution and pooling layers. To enable the flow of long-range information across the manifold $\partial\Omega$, we opt for an encoder-decoder architecture with three pooling levels and \quote{copy \& concatenate} connections between corresponding layers in the contracting and expanding pathway.
To prevent vanishing gradients, we use residual blocks
consisting of two convolution layers and a skip connection. We use ReLU activation functions and employ batch normalisation before each activation.

\subsection{Convolution layer}


We define signals $f \colon \V \to \R^c$ with channel size $c$. For ease of notation, we compactly denote the set of all fields mapping from $\V$ to $\R^c$ as $\X(\V, \R^c)$ so that we can write $f \in \X(\V, \R^c)$. As a central building block of our neural network, we define convolution layers on $\M$ via message passing~\cite{GilmerSchoenholz2017}. Let $c_i$ and $c_{i + 1}$ denote the channel size before and after the layer.
\begin{equation*}
    (\phi * f) \colon
    \begin{cases}
    \X(\V, \R^{c_i}) \to \X(\V, \R^{c_{i + 1}}) \\
    f \mapsto \gamma(\phi(p, f))  
    \hspace{0.6cm} \forall p \in \V.
    \end{cases}
\end{equation*}
The messages $\phi$ aggregate information from the neighbourhood $\B_r(p) \cap \V$, where $\B_r(p)$ consists of all vertices that are contained in a ball with radius $r$ around $p \in \V$. The update function $\gamma$ creates the signal update from these messages. Alternatively, the neighbourhood could be defined by a 1-ring neighbourhood on the mesh $\M$ or by a geodesic ball on the manifold $\partial\Omega$. Our definition is an approximation to these options that is robust to varying mesh resolutions and scalable to large meshes. We construct convolution layers with kernel $K \colon \V \times \V \to \R^{c_i \times c_{i + 1}}$ by choosing the messages
\begin{equation}
    \phi(p, f) \coloneqq \sum\limits_{q \in \B_r(p) \cap \V} K(p, q) \rho(p, q) f(q)
    \label{eq:conv}
\end{equation}

We refer to a neural network containing the aforementioned convolution layer as mesh-based GCN with the following rationale. The neighbourhood of a mesh vertex induces a set of graph edges $\E$ by connecting $p$ to all $q \in \B_r(p) \cap \V$.
With this \quote{latent} graph structure $(\V, \E)$ we can make use of efficiently implemented graph deep-learning libraries (like PyG)
to realise our layers. Additionally, this GCN can be \textbf{mesh-based} by explicitly incorporating face information in the message passing $\phi = \phi^{(\V, \F)}$.

We distinguish between \textbf{isotropic} and \textbf{anisotropic} convolution layers based on kernel $K(p, q)$ and aggregation matrix $\rho(p, q) \colon \V \times \V \to \R^{c_i \times c_i}$. Intuitively, isotropic convolution filters process all signals mapped to the surrounding vertices in a neighbourhood in the same manner, while anisotropic filters process them distinctly.
\begin{definition}[Anisotropy]
	We call bivariable functions $G \colon p, q \mapsto G(p, q)$ with $p \in \V$ and $q \in \B_r(p) \cap \V$ isotropic, if $G(p, q)$ is constant in $q$. We call $G$ anisotropic, if $G(p, q)$ is not constant in $q$ for all $q \in \B_r(p) \cap \V$. Consequently, we call a layer anisotropic, if it contains any anisotropic function.
\end{definition}

\subsubsection{Gauge-equivariant mesh convolution}

Defining general anisotropic kernels $K(p, q)$ on meshes is difficult due to the lack of a local canonical orientation on the mesh: there is no obvious choice of reference vertex $q \in \B_r \cap \V$ in the filter support that canonically orients the local filter at $p$ for all $p \in \V$. To address this, we implement anisotropic kernels using gauge-equivariant mesh (GEM) convolution~\cite{HaanWeiler2021}. The idea behind GEM convolution is to recognise that possible kernel orientations are related by group actions of the symmetry group of planar rotations $\SO(2)$ and use this insight to spatially orient kernels \quote{along} its group elements.

To achieve this, the signal $f \in \X(\V, \R^c)$ is composed of a linear combination of irreducible representations (\quote{irreps}) of the symmetry group $\SO(2)$, resulting in so-called $\SO(2)$ features.
We can then choose an invertible parallel transport matrix
\begin{equation*}
    \rho(p, q) = \rho(p, q)_{(\V, \F)}
\end{equation*}
composed of group action representations that can rotate signals $f$ using mesh information. Specifically, the tangential plane at each vertex can be determined from the surrounding triangles and geodesic shortest paths between vertices can be found from adjacent faces~\cite{SharpCrane2020}. Parallel transport refers to transporting signals along the manifold $\partial\Omega$ while maintaining a fixed angle to the shortest geodesic curve. It provides a unique and thus canonical transformation that allows linearly combining vector fields $f \in \X(\V, \R^c)$ at a vertex $p \in \V$ on the mesh. This is required for our notion of convolution Eq.~(\ref{eq:conv}).

On 2D manifolds $\partial\Omega$ embedded in 3D Euclidean space, picking a kernel orientation amounts to picking a locally tangential coordinate system (\quote{gauge}). This choice can, on general manifolds, only be made arbitrarily. To prevent this to arbitrarily affect the outcome of the convolution, GEM convolution imposes an \textit{equivariance} relation between layer input and output. Let $P$ and $P'$ be representations of the same (linear) gauge transformation that rotates the feature vector. GEM convolution requires message passing Eq.~(\ref{eq:conv}) to be equivariant under such transformations. Since all other variables in Eq.~(\ref{eq:conv}) are fixed, this imposes a linear constraint on the kernel $K(p, q)$ with solutions
\begin{equation*}
    \{K(p, q) \hspace{0.2cm} | \hspace{0.2cm} P' \phi(p, f) \overset{!}{\equiv} \phi(p, Pf)\}
\end{equation*}
A detailed derivation can be found in \cite{HaanWeiler2021}.

\subsection{Pooling}\label{subsec:pooling}

Hemodynamics are characterised by long-range interactions across the artery wall $\partial\Omega$ and the lumen $\Omega$. Capturing these by stacking convolution layers, i.e. linearly increasing the receptive field, becomes infeasible for large and finely discretised surfaces. In contrast, pooling layers can exponentially increase the network's receptive field. Here, we use the mesh's \quote{latent} computation graph $(\V, \E)$ to implement pooling. Similar to the procedure used by Wiersma et al.~\cite{WiersmaEisemann2020}, we sample a hierarchy of vertex subsets $(\V = \V_0) \supset \V_1 \supset \dots \supset \V_n$ and construct according $r$-radius graph edges $\E_i$ encoding the filter support $\B_{r_i}(p) \cap \V_i$ for all $p \in \V_i$. Additionally, we find disjoint partitions of clusters
\begin{equation*}
    \bigcup\limits_{p \in \V_{i + 1}} C(p) = \V_i, \hspace{1.0cm} \bigcap\limits_{p \in \V_{i + 1}} C(p) = \emptyset
\end{equation*}
that relate fine-scale vertices to exactly one coarse-scale vertex. This can be done with $k$-nearest neighbours ($k = 1$) by finding for each $p \in \V_i$ the nearest vertex in $\V_{i + 1}$. Using these, a pooling operator can be defined as
\begin{equation*}
    \psi_\text{pool} \colon
    \begin{cases}
    \X(\V_i, \R^c) \to \X(\V_{i + 1}, \R^c) \\
    f \mapsto \frac{1}{|C(p)|} \sum\limits_{q \in C(p)} \rho(p, q) f(q)
    \hspace{0.4cm} \forall p \in \V_{i + 1}
    \end{cases}
\end{equation*}
We implement unpooling by simply transporting signals $f$ back to their respective cluster locations:
\begin{equation*}
    \psi_\text{unpool} \colon
    \begin{cases}
    \X(\V_{i + 1}, \R^c) \to \X(\V_i, \R^c) \\
    f \mapsto \rho^{-1}(C^{-1}(p), p) f(C^{-1}(p))
    \hspace{0.4cm} \forall p \in \V_i
    \end{cases}
\end{equation*}

\subsection{Input features}\label{subsec:features}

We construct input features $f^\text{in} \colon \V \to \R^{c_\text{in}}$ with $c_\text{in}$ channels that describe the local shape of $\partial\Omega$ as well as global properties and are computed from the mesh $\M$. In particular, we compute a surface normal for each vertex $p \in V$ from adjacent mesh faces. We then construct three matrices that describe the local neighbourhood $q \in \B_r(p) \cap \V$ by, for each neighbour $q$, taking the outer products of
\begin{itemize}
    \item the vector from $p$ to $q$ with itself,
    \item the surface normal at $q$ with itself, and
    \item the vector from $p$ to $q$ with the surface normal at $q$
\end{itemize}
For each of the three resulting sets of ($3 \times 3$)-matrices, we take the average over the neighbourhood. Two of these matrices are symmetric by construction, so we can drop entries without losing information. The radius $r$ of the local neighbourhood balls is a hyperparameter and must be chosen based on the structure of the input meshes, so that no neighbourhood is disconnected, i.e. consists of a single vertex. We chose the same radius that is used to construct the mesh's \quote{latent} computation graph $(\V_0, \E_0)$.

The motivation behind these input features is that they define meaningful local surface descriptors that are not $\SO(2)$-invariant, a precursor to employing GEM convolution~\cite{HaanWeiler2021}. In contrast, the vanilla surface normal would simply be constant in any coordinate system induced by the surface normal. Since the surface normal describes the local surface (orientation) in an infinitesimally small neighbourhood $\B_{r \to 0}(p)$, i.e. the precise local curvature of the artery wall $\partial\Omega$, it is the preferred input feature for conventional message passing formulations.

We can extend the per-vertex features with any scalar or vector field. Since we assume that hemodynamics depend on flow direction, we append the shortest geodesic distance from each vertex $p$ to the inflow surface, which we compute with the vector heat method~\cite{SharpSoliman2019}. Moreover, we add global parameters such as blood-flow boundary conditions as a constant scalar field over the vertices.

\subsection{Network output}\label{subsec:output}

We predict vector-valued hemodynamic quantities arising from transient, pulsatile flow by discretising a full cardiac cycle at $T$ points in time and let our neural network output a vector field $f^\text{out} \in \X(\V, \R^{3 T})$. Alternatively, we can predict hemodynamic fields under steady flow by setting $T = 1$.

\subsection{$\SE(3)$ equivariance}

We model hemodynamics without the influence of gravity. Therefore, rigid rotation (or translation) of the domain should  have no influence on the magnitude of the flow quantities and only change their direction. More precisely, our problem exhibits equivariance under $\SE(3)$ transformation. Inducing this symmetry in our neural network makes it oblivious to particular transformations which reduces the problem's complexity. We do so in the form of GEM convolution.

\begin{proposition}\label{prop:equivariance}
    (Informal) Composition of rotation-equivariant and translation-invariant input features with a gauge-equivariant mesh (graph) convolutional neural network (GEM-GCN) is end-to-end $\SE(3)$-equivariant. (proof in Appendix~\ref{app:proof})
\end{proposition}

GEM convolution layers define message passing intrinsically on the mesh $\M$ without dependence on the embedding in the ambient space, such as Euclidean vertex coordinates. $\SO(2)$ features can be expressed in ambient coordinates, which is done at the network output.
Since tangential planes by definition rotate with the geometric model of the artery, the GEM convolution operator $(K * f)$ preserves $\SE(3)$ equivariance if the tangential input features move along with the surface.

Our input features $f^\text{in}$ are equivariant under rotation and invariant under translation of the mesh $\M$ by construction. Furthermore, our pooling and unpooling operators $\psi_\text{pool}$ and $\psi_\text{unpool}$ preserve $\SE(3)$ equivariance because they do not depend on the embedding of $\M$ in ambient space. Consequently, neural networks composed entirely of GEM convolution and pooling layers yield an end-to-end $\SE(3)$-equivariant operator together with our input features $f^\text{in}$.


\subsection{Baseline models}\label{subsec:baselines}

\input{figures/conv}

We perform ablation studies to investigate the influence of the anisotropic aggregation matrix $\rho(p, q)$ and the anisotropic kernel $K(p, q)$ on prediction accuracy. To this end, we define two additional types of convolution (Figure~\ref{fig:conv}): one fully isotropic and one with a learned anisotropic aggregation matrix. Additionally, we compare our method to another baseline model, PointNet\texttt{++}~\cite{QiYi2017},
a point cloud method without explicit convolution kernels.

\subsubsection{Isotropic convolution}

We construct purely isotropic convolution by choosing
\begin{align*}
    \rho(p, q) &= \rho \coloneqq \I \\
    K(p, q) &= K(p) \coloneqq \frac{1}{|\B_r(p) \cap \V |} W
\end{align*}
in Eq.~(\ref{eq:conv}) where $\I$ is the identity matrix and $W \in \R^{c_i \times c_{i + 1}}$ are trainable weights.

\subsubsection{Attention-scaled convolution}

We construct anisotropic convolution with an isotropic kernel via a learned neighbourhood-attention mechanism
by choosing:
\begin{align*}
    \rho(p, q) &\coloneqq \sigma((f(q) - f(p)) \cdot w) \I \\
    K(p, q) &= K(p) \coloneqq \frac{1}{|\B_r(p) \cap \V|} W
\end{align*}
in Eq.~(\ref{eq:conv}) where $\sigma(\cdot)$ is the element-wise softmax activation and $W \in \R^{c_i \times c_{i + 1}}$ as well as $w \in \R^{c_i}$ are trainable weights. This is equivalent to a graph attention layer~\cite{VelickovicCucurull2018} with separate weights and no LeakyReLU activation in the attention mechanism. Note that here, the message passing is not mesh-based and only depends on the vertices: $\phi = \phi^\V$.

In our definition of pooling in Sec.~\ref{subsec:pooling} we require the inverse of $\rho$ for the unpooling step. Since for attention-scaled convolution, $\rho$ may be ill-conditioned with diagonal elements close to zero, we fall back to using $\I$ for pooling.

\subsubsection{PointNet\texttt{++}}

We compare our method to PointNet\texttt{++}~\cite{QiYi2017},
a popular point cloud method consisting of message passing layers that redefine Eq.~\ref{eq:conv} by
\begin{equation}
    \phi_k = \max\limits_{q \in \B_r(p) \cap \V} \Theta_k(f(q), v_{p \to q})
    \label{eq:pointnet}
\end{equation}
where $k \leq c_{i + 1}$ denotes the $k$-th component, $v_{p \to q}$ the Euclidean vector pointing from $p$ to $q$, and $\Theta \colon \R^{c_i} \times \R^3 \to \R^{c_{i+1}}$ an MLP of arbitrary depth. PointNet\texttt{++} uses sampling and grouping operations that hierarchically sub-sample the graph vertices in the contracting pathway and interpolate in the expanding pathway. Note that, for PointNet\texttt{++}, choosing the same pooling architecture as for the kernel-based GCNs does not lead to the same level of accuracy, since the convolution paradigms are fundamentally different. Thus, we lay out PointNet\texttt{++} separately, to achieve the best possible performance.

\subsection{Quantitative evaluation}

Quantitative results for WSS estimation are reported in terms of mean absolute error of the elements of $\triangle$, normalised by the maximum ground truth magnitude across the test split (\quote{NMAE}) and approximation error $\varepsilon \coloneqq \Vert \triangle \Vert_2 / \Vert L \Vert_2$. $\triangle$ is a vector whose elements are vertex-wise $\L^2$-normed differences between the network output $f^\text{out} \in \X(\V, \R^{c_\text{out}})$ and ground truth label $l \in \X(\V, \R^{c_\text{out}})$ so that the $i$-th element of vector $\triangle_i = \Vert f^\text{out}(p^i) - l(p^i) \Vert_2$ and $L_i = \Vert l(p^i) \Vert_2$ for $p^i \in \V$. Additionally, we report the maximum and mean vertex-wise difference, i.e. $\triangle^\text{max} = \max \{\triangle_i\}_i$ and $\triangle^\text{mean} = (\sum_i \triangle_i) / |\V|$ as well as the mean of the label statistics $\max \{L_i\}_i$ and $\text{median} \{L_i\}_i$ over the test set for scale.

%% file: figures/overview.tex
\begin{figure*}
    \centering
    \includegraphics[width=\textwidth]{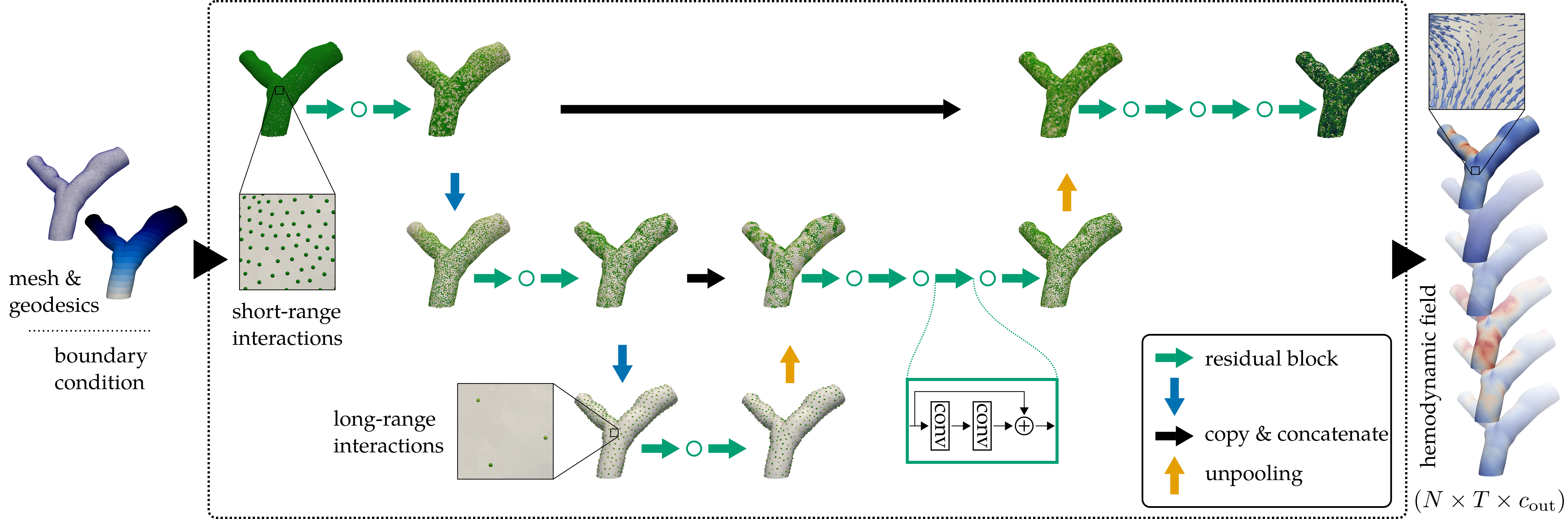}
    \caption{\textbf{Network architecture.} Our mesh-based GCN outputs time-discretised, pulsatile hemodynamic fields $f^\text{out} \colon \V \to \R^{T \times c_\text{out}}$, where $|\V| = N$, subject to a (scalar) coronary blood flow parameter, given an input consisting of artery-wall mesh and vertex-wise geodesic distance to the artery inlet. A large receptive field is efficiently obtained using a three-level pooling scheme. To enable deep networks, we employ residual blocks consisting of two convolution modules and skip connection. The per-vertex colour of the signal before and after residual blocks corresponds to the scalar activation mapped to the vertices.}
    \label{fig:overview}
\end{figure*}

%% file: figures/conv.tex
\begin{figure}
    \centering
    \includegraphics[width=\columnwidth]{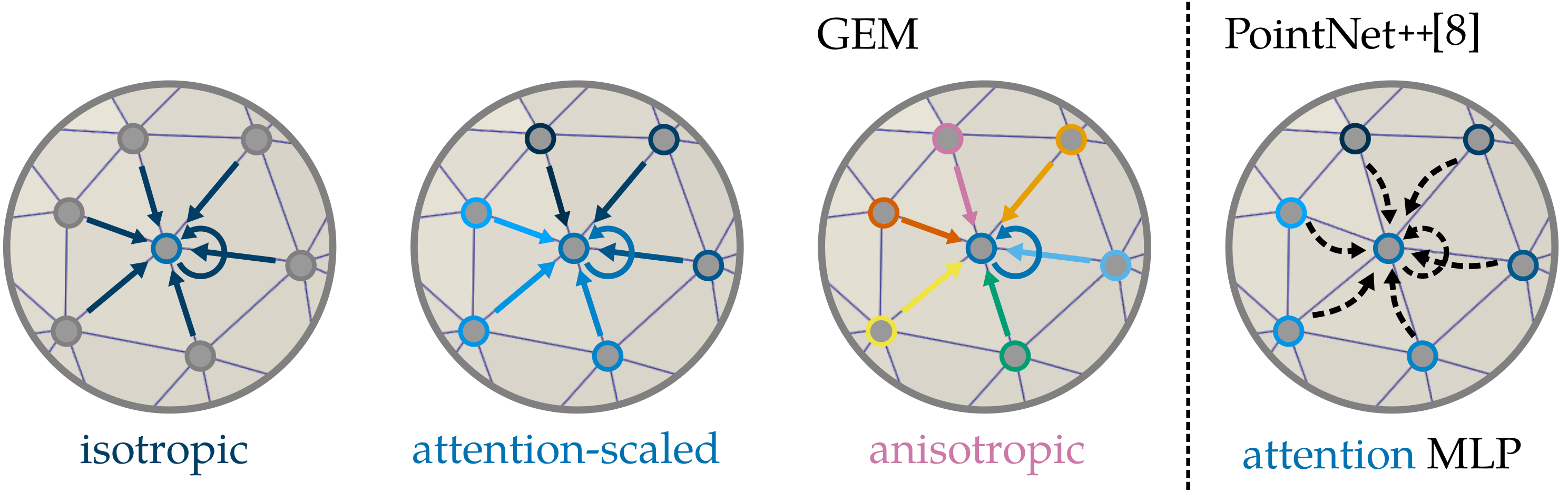}
    \caption{\textbf{Filter comparison.} Isotropic, attention-scaled, and GEM convolution use kernels, in comparison to PointNet\texttt{++} message passing. While attention-scaled convolution and PointNet\texttt{++} both learn to distinguish neighbouring vertices through an attention mechanism, GEM convolution is equipped with a notion of direction.}
    \label{fig:conv}
\end{figure}

%% file: content/experiments.tex
\section{Experiments and results}\label{sec:experiments}

We evaluate to what extent GEM-GCN can predict directional wall shear stress on the artery models described in Sec.~\ref{sec:data}. All datasets are split 80:10:10 into training, validation, and test splits, respectively. Network width and depth are set so that each neural network has around \num{1.02e6} trainable weights. All neural networks are trained by stochastic $\L^1$-loss regression using an Adam optimiser
with batches of 12 samples and a learning rate of \num{1e-3} on ground truth values obtained by CFD. All experiments are run on NVIDIA A40 (48 GB) GPUs. Parallelisation over two GPUs was necessary to fit batches of 12 bifurcating artery models into memory. Inference for a previously unseen artery wall takes less than 5 s including geometric pre-processing. Our open-source implementation in PyTorch
and PyG
using the vector heat method~\cite{SharpSoliman2019} can be found online.\footnote{\url{github.com/sukjulian/coronary-mesh-convolution}}

\subsection{Steady-flow WSS estimation}\label{subsec:steady}

\input{figures/steady}
\input{tables/metrics}

We train GEM-GCN as well as the isotropic GCN (IsoGCN), the attention-scaled GCN (AttGCN), and PointNet\texttt{++} (Sec.~\ref{subsec:baselines}) to perform WSS estimation in the steady-flow single and bifurcating artery datasets. Fig.~\ref{fig:steady} shows examples of directional WSS prediction by GEM-GCN in a single and a bifurcating artery. The examples suggest that there is good agreement between ground truth and prediction. In particular, \revision{WSS stemming from local flow vortices is captured well} in the single artery model. The quantitative results in Table~\ref{tab:metrics} show that GEM-GCN strictly outperforms IsoGCN and AttGCN on both the single and the bifurcating artery dataset. Moreover, the learned anisotropic convolution filters used in AttGCN achieve better performance than the isotropic filters used in IsoGCN. GEM-GCN and {PointNet\texttt{++}} perform similarly in accuracy on the bifurcating artery dataset while GEM-GCN performs marginally better on the single arteries.

\input{plots/efficiency}

We evaluate how the amount of training data affects performance of GEM-GCN, as well as PointNet++ for comparison. Fig.~\ref{plt:efficiency} shows mean approximation error $\varepsilon_\text{mean}$ as a function of the number of training samples. For each training set size, GEM-GCN is trained from scratch on the single artery dataset, for a number of epochs chosen so that it receives ca. 10,000 gradient-descent updates. Since PointNet\texttt{++} requires more epochs to converge we train it for 80,000 gradient-descent updates for comparison. The results in Fig.~\ref{plt:efficiency} indicate that both architectures can reach good accuracy with ca. 1000 training samples.

\subsection{$\SO(3)$ equivariance}\label{subsec:equivariance}

GEM-GCN only depends on relative vertex features and is trivially invariant to translation. To empirically verify $\SO(3)$ equivariance of GEM-GCN, we perform predictions on randomly rotated test samples. For this we use the neural network trained on the original, canonically oriented samples. The results in Table~\ref{tab:metrics} show that rotation does indeed not affect performance of GEM-GCN. All quantitative metrics are nearly identical to those on the non-rotated samples up until numerical errors originating from discretisation of the kernels and activation function~\cite{HaanWeiler2021}. In contrast, results show that for PointNet\texttt{++} (the best-performing baseline model) rotation of test samples drastically reduces prediction accuracy: performance drops from a mean NMAE of 0.5~\% to 10.1~\% for the single and 0.6~\% to 7.8~\% for the bifurcating artery dataset, respectively. This is expected as PointNet\texttt{++} -- like previously published models \cite{LiangMao2020,LiWang2021,MoralesFerezMill2021} -- depends on the embedding of the mesh vertices in Euclidean space.

In order to make PointNet\texttt{++} account for differently rotated samples, we re-train it with data augmentation by batch-wise, randomly sampling rotation matrices and applying them to the training samples. This is a common strategy for methods that lack rotation equivariance.  Results show that training with this augmentation approximately recovers PointNet\texttt{++}'s accuracy to 0.7~\% and 0.6~\% mean NMAE for single and bifurcating arteries, respectively. This is slightly lower than before for the single arteries. However, training time until convergence is roughly 1.5 times longer, going from 20:48 [h] to 31:29 [h] and 35:01 [h] to 57:16 [h] for single and bifurcating arteries, respectively.

\subsection{Pulsatile-flow WSS estimation}

\input{plots/temporal}

We train GEM-GCN for pulsatile-flow WSS estimation in single arteries with the modifications described in Sec.~\ref{subsec:output}. 
In these experiments, WSS is dependent on both space and time. Therefore, we present estimation accuracy as time-dependent distributions in Fig.~\ref{plt:temporal}. The pulsatile-flow NMAE over time is comparable to the steady-flow NMAE, suggesting generally accurate predictions. However, the pulsatile-flow NMAE depends on the maximum WSS, which fluctuates over the cardiac cycle. As a consequence, the NMAE fluctuates as well and follows the pattern of the maximum WSS (indicated in yellow).

\subsection{Incorporating boundary conditions}\label{subsec:bct}

We re-train GEM-GCN on the dataset of pulsatile-flow WSS in single and bifurcating arteries, subject to varying coronary blood flow boundary conditions. \revision{The boundary conditions are passed as average flow rate of a scaled template waveform (Fig.~\ref{fig:waveform}).} We investigate interpolation between and extrapolation to different boundary conditions outside the limits of the training distribution: As described in Sec.~\ref{sec:data}, values in $[1.87, 4.36]$ $\frac{\text{ml}}{\text{s}}$ are contained in the training data and neural-network predictions subject to boundary conditions within this domain require interpolation. Values in $[0.63, 1.87]$ $\frac{\text{ml}}{\text{s}}$ and $[4.36, 5.61]$ $\frac{\text{ml}}{\text{s}}$ require extrapolation, as GEM-GCN is not trained on simulations subject to these inflow values. However, GEM-GCN \revision{will produce a} prediction based on \revision{any arbitrary flow rate}. Here, we restrict our analysis to a discrete set of boundary conditions from a continuous range for which we have performed CFD simulation (Sec.~\ref{sec:data}).

\input{plots/extrapolation}

Fig.~\ref{plt:extrapolation} quantifies the prediction error for varying boundary conditions in two ways: First, we plot (mean) NMAE over coronary blood flow from which we observe the following: within the training range, the infimum of the NMAE displays a linear dependence on the boundary condition. The NMAE values corresponding to boundary conditions higher than this training range stay below this slope, while the NMAE values corresponding to lower values go above it. Second, we show a Bland-Altman plot comparing neural-network prediction and ground-truth reference. This plot shows that GEM-GCN overestimates WSS for low average magnitude and underestimates WSS for high average magnitude. A large amount of data points corresponding to extrapolation fall within the upper and lower bounds of the distribution of interpolated data points. From these two plots we conclude that GEM-GCN extrapolates to some extent to boundary condition values higher than those in the ground-truth distribution.

\subsection{Sensitivity to remeshing}

\input{figures/remeshing}

Recent works suggest that mesh neural networks might overfit to mesh connectivity~\cite{SharpAttaiki2022}. For the problem of estimating hemodynamics on polygonal surface meshes this means that predictions are not independent of the sampling of vertex positions on the underlying manifold. To investigate the susceptibility of our models to overfitting, we let the trained GEM-GCN and PointNet\texttt{++} networks described in Sec.~\ref{subsec:steady} estimate WSS fields on three kinds of remeshed versions of the same surface $\partial\Omega$ of a sample from the test set of the single arteries:
\begin{enumerate}
    \item We randomly sample vertices from $\partial\Omega$ and apply Poisson surface reconstruction, followed by an isotropic meshing procedure. This relaxes the mesh refinement around the stenoses and leads to approximately \textbf{equidistant} vertex spacing.
    \item We globally \textbf{refine} the original mesh $\M$ so faces $\F$ have smaller edge lengths, while maintaining proportionally higher resolution around the stenoses.
    \item We randomly sample mesh vertices from $\partial\Omega$, completely \textbf{randomising} vertex placement beyond refinement or coarsening. GEM-GCN extracts mesh information from the vertices and corresponding surface normals, which are well-defined here. Thus, we can do without an explicit mesh in this particular case.
\end{enumerate}
The results in Fig.~\ref{fig:remeshing} suggest that GEM-GCN is still able to identify regions of interest on the surface $\partial\Omega$: in the equidistant mesh, it predicts high WSS magnitude in the stenosed area even with different mesh connectivity. However, GEM-GCN does overfit, to some extent, to mesh connectivity: regions of high vertex density, especially in the refined mesh, are predicted to have high WSS magnitude and vice versa. This might be because the training data has higher resolution around stenoses and WSS values are typically highest in stenotic regions. Thus, the network learns that high resolution corresponds to high WSS. The predictions on randomly sampled vertices show artifacts of this behaviour in the form of arbitrary peaks, caused by high local vertex density. This conditioning on resolution may be due to the aggregation scheme (see Equation~\eqref{eq:conv}) used by GEM convolution: the filters sum over the vertex neighbourhoods, as opposed to e.g. taking the maximum. PointNet\text{++} seems more robust to remeshing and random surface sampling, perhaps due to its maximum-aggregation (see Equation~\revision{\eqref{eq:pointnet}}) scheme.

\subsection{Generalisation to real-life patient data}

\input{figures/real}

While we develop and evaluate our method on synthetic data, clinical application would be on anatomies extracted from individual patients. To assess generalisation to such data, we use the same GEM-GCN trained on the bifurcating arteries from Sec.~\ref{subsec:steady} and let it predict WSS in a left main coronary bifurcation geometry extracted from a cardiac CT angiography scan~\cite{WolterinkLeiner2019}.
We simulate blood flow with the same boundary conditions as in Sec.~\ref{sec:data} to obtain ground-truth WSS which takes ca. 30 min. Fig.~\ref{fig:real} shows the ground truth and estimated WSS vectors. As previously, prediction and geometric pre-processing take less than 5 s. Even though GEM-GCN is trained exclusively on synthetic arteries, it produces a qualitatively plausible prediction. \revision{More precisely, the directions of the WSS vectors agree well between prediction and ground truth (mean cosine similarity 0.97).} However, there is a considerable quantitative error \revision{($\text{NMAE}_\text{mean}$ 87.4~\%, $\varepsilon_\text{mean}$ 12.1~\%)} which can be explained by the highly nonlinear dependence of blood flow on lumen wall shape: even small differences in morphology between the synthetic and real-life arteries can influence hemodynamics to an extent that cannot be easily extrapolated by GEM-GCN. Nevertheless, Fig.~\ref{fig:real} suggests that GEM-GCN is able to qualitatively transfer the relation between local surface curvature and WSS.

%% file: figures/steady.tex
\begin{figure}
    \centering
    \includegraphics[width=\columnwidth]{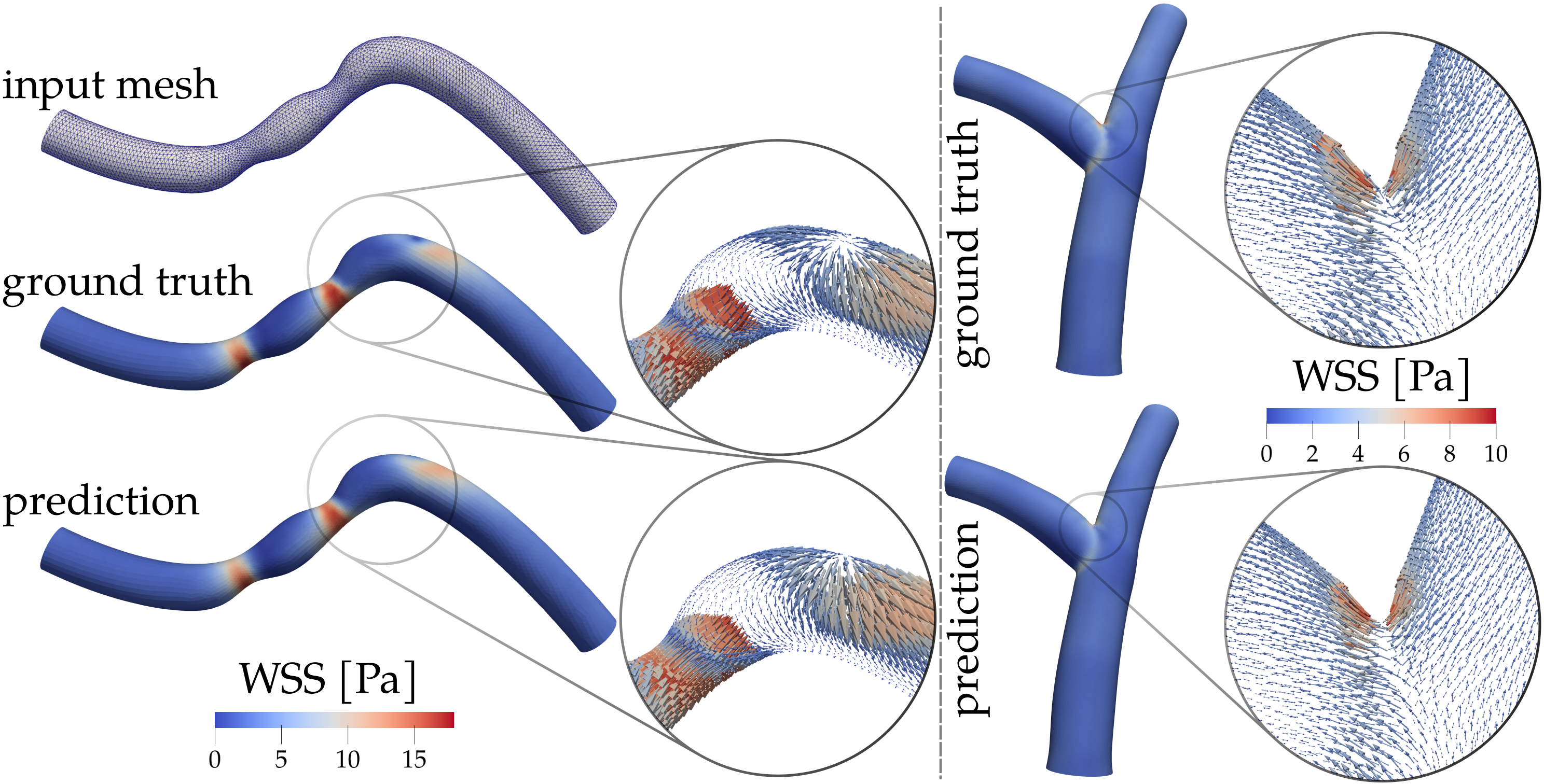}
    \caption{\textbf{Steady-flow WSS estimation} of GEM-GCN on arteries of the held-out test splits of the single (left) and bifurcation artery (right) datasets.}
    \label{fig:steady}
\end{figure}

%% file: tables/metrics.tex
\begin{table*}
	\begin{center}
    	\caption{\textbf{Quantitative evaluation} of prediction error for steady-flow WSS on synthetic single and bifurcating coronary arteries. The columns list the mean, median, and 75th percentile of NMAE, approximation error $\varepsilon$, maximum absolute error $\Delta^\text{max}$, and mean absolute error $\Delta^\text{mean}$ over the held-out test splits. Maximum and median WSS magnitude per dataset are indicated as $L_\text{max}$ and $L_\text{median}$, respectively. We additionally evaluate PointNet\texttt{++} and GEM-GCN on randomly 3D-rotated test samples with previous training on canonically oriented samples ($^\dagger$). In the rotated case we additionally present accuracy metrics for PointNet\texttt{++} for training on rotationally augmented data ($^\ddagger$).}
    	\resizebox{\textwidth}{!}{
    		\renewcommand{\arraystretch}{1.3}
    		\begin{tabular}{@{}ccccccccccccccccccccc@{}}
    			\toprule
    			&&&& \multicolumn{3}{c}{NMAE [\%]} & \phantom{abc} & \multicolumn{3}{c}{$\varepsilon$ [\%]} & \phantom{abc} & \multicolumn{3}{c}{$\triangle^\text{max}$ [Pa]} & \phantom{abc} & \multicolumn{3}{c}{$\triangle^\text{mean}$ [Pa]} &&\\
    			\cmidrule{5-7} \cmidrule{9-11} \cmidrule{13-15} \cmidrule{17-19}
    			&&&& mean & median & $75$th && mean & median & $75$th && mean & median & $75$th && mean & median & $75$th &&\\
    			\midrule
    			\multirow{7}{1.6cm}{\textbf{Single\\ arteries}} &
    			\multirow{4}{1.0cm}{\textbf{oriented}}
    			& IsoGCN && 0.9 & 0.9 & 1.2 && 15.7 & 15.2 & 19.3 && 5.93 & 5.88 & 7.96 && 0.45 & 0.45 & 0.60 && \multirow{8}{2.7cm}{$L_\text{max}$ = 22.53 [Pa]\\ $L_\text{median}$ = 2.07 [Pa]}\\
    			&& AttGCN && 0.6 & 0.6 & 0.8 && 10.1 & 9.7 & 11.9 && 4.33 & 3.78 & 6.38 && 0.31 & 0.30 & 0.41 &&\\
    			&& PointNet\texttt{++} && \textbf{0.5} & \textbf{0.4} & 0.7 && 8.6 & 8.2 & 11.0 && 4.67 & 3.87 & 7.14 && 0.25 & \textbf{0.21} & 0.34 &&\\
    			&& GEM-GCN && \textbf{0.5} & \textbf{0.4} & \textbf{0.6} && \textbf{7.8} & \textbf{7.6} & \textbf{9.1} && \textbf{4.10} & \textbf{3.55} & \textbf{6.13} && \textbf{0.23} & 0.23 & \textbf{0.31} &&\\
    			\cmidrule{2-19}
    			& \multirow{3}{1.0cm}{\textbf{rotated}}
    			& PointNet\texttt{++}$^\dagger$ && 10.1 & 10.0 & 11.9 && 154.4 & 141.1 & 180.6 && 31.18 & 28.7 & 41.73 && 5.14 & 5.09 & 6.04 &&\\
    			&& PointNet\texttt{++}$^\ddagger$ && 0.7 & 0.6 & 1.0 && 12.3 & 11.4 & 15.9 && 6.17 & 5.41 & 8.97 && 0.36 & 0.32 & 0.49 &&\\
    			&& GEM-GCN$^\dagger$ && \textbf{0.5} & \textbf{0.4} & \textbf{0.6} && \textbf{7.7} & \textbf{7.5} & \textbf{9.2} && \textbf{4.10} & \textbf{3.50} & \textbf{5.79} && \textbf{0.23} & \textbf{0.22} & \textbf{0.31} &&\\
    			\midrule
    			\multirow{7}{1.6cm}{\textbf{Bifurcating\\ arteries}} &
    			\multirow{4}{1.0cm}{\textbf{oriented}}
    			& IsoGCN && 1.0 & 0.9 & 1.0 && 16.9 & 15.3 & 17.4 && 3.64 & 3.34 & 4.24 && 0.19 & 0.17 & 0.20 && \multirow{8}{2.7cm}{$L_\text{max}$ = 7.16 [Pa]\\ $L_\text{median}$ = 1.37 [Pa]}\\
    			&& AttGCN && 0.7 & 0.6 & 0.7 && 12.6 & 11.3 & 13.0 && 3.50 & 3.34 & 4.07 && 0.14 & 0.12 & 0.14 &&\\
    			&& PointNet\texttt{++} && \textbf{0.6} & \textbf{0.5} & \textbf{0.6} && \textbf{11.2} & \textbf{10.5} & \textbf{12.1} && \textbf{3.29} & \textbf{2.96} & 4.01 && \textbf{0.12} & \textbf{0.10} & \textbf{0.13} &&\\
    			&& GEM-GCN && \textbf{0.6} & 0.6 & 0.7 && 11.9 & 11.3 & 13.0 && 3.38 & 3.25 & \textbf{3.92} && 0.13 & 0.11 & \textbf{0.13} &&\\
    			\cmidrule{2-19}
    			& \multirow{3}{1.0cm}{\textbf{rotated}}
    			& PointNet\texttt{++}$^\dagger$ && 7.8 & 7.6 & 11.0 && 114.6 & 124.7 & 153.9 && 7.81 & 7.98 & 9.52 && 1.56 & 1.52 & 2.18 &&\\
    			&& PointNet\texttt{++}$^\ddagger$ && \textbf{0.6} & \textbf{0.6} & \textbf{0.7} && 12.3 & 11.5 & 13.5 && 3.48 & 3.28 & 4.01 && \textbf{0.13} & \textbf{0.11} & \textbf{0.14} &&\\
    			&& GEM-GCN$^\dagger$ && \textbf{0.6} & \textbf{0.6} & \textbf{0.7} && \textbf{12.1} & \textbf{11.3} & \textbf{13.2} && \textbf{3.42} & \textbf{3.25} & \textbf{3.91} && \textbf{0.13} & 0.12 & \textbf{0.14} &&\\
    			\bottomrule
    			&& \multicolumn{19}{l}{$^\dagger$ trained on canonically oriented samples}\\
    			&& \multicolumn{19}{l}{$^\ddagger$ trained under data augmentation (random rotation in 3D)}\\
    		\end{tabular}
    	}	
    	\label{tab:metrics}
	\end{center}
\end{table*}

%% file: plots/efficiency.tex
\begin{figure}
    \centering
    \includegraphics[width=\columnwidth]{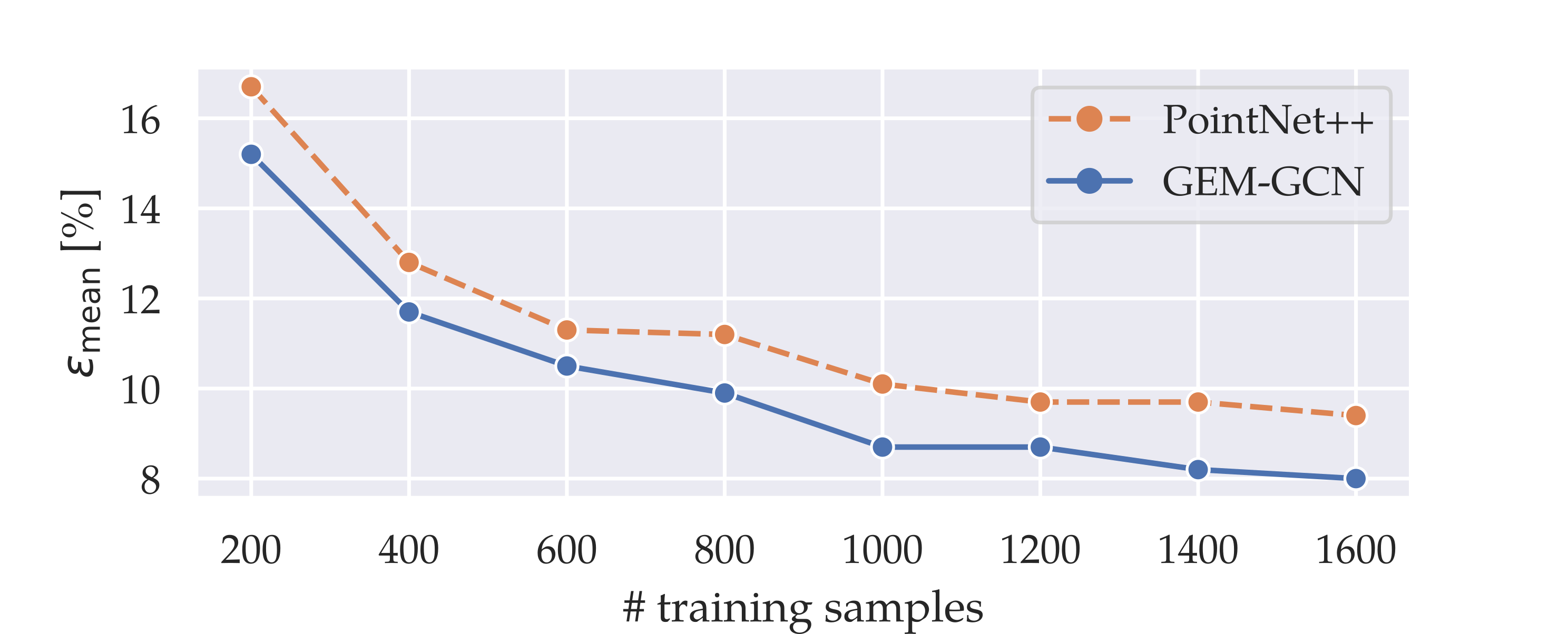}
    \caption{\textbf{Mean approximation error} $\varepsilon_\text{mean}$ over the test split for different training set sizes on the steady-flow single-artery dataset. GEM-GCN weights are updated for ca. 10,000 iterations, PointNet++ weights for ca. 80,000 iterations.}
    \label{plt:efficiency}
\end{figure}

%% file: plots/temporal.tex
\begin{figure}
    \centering
    \includegraphics[width=\columnwidth]{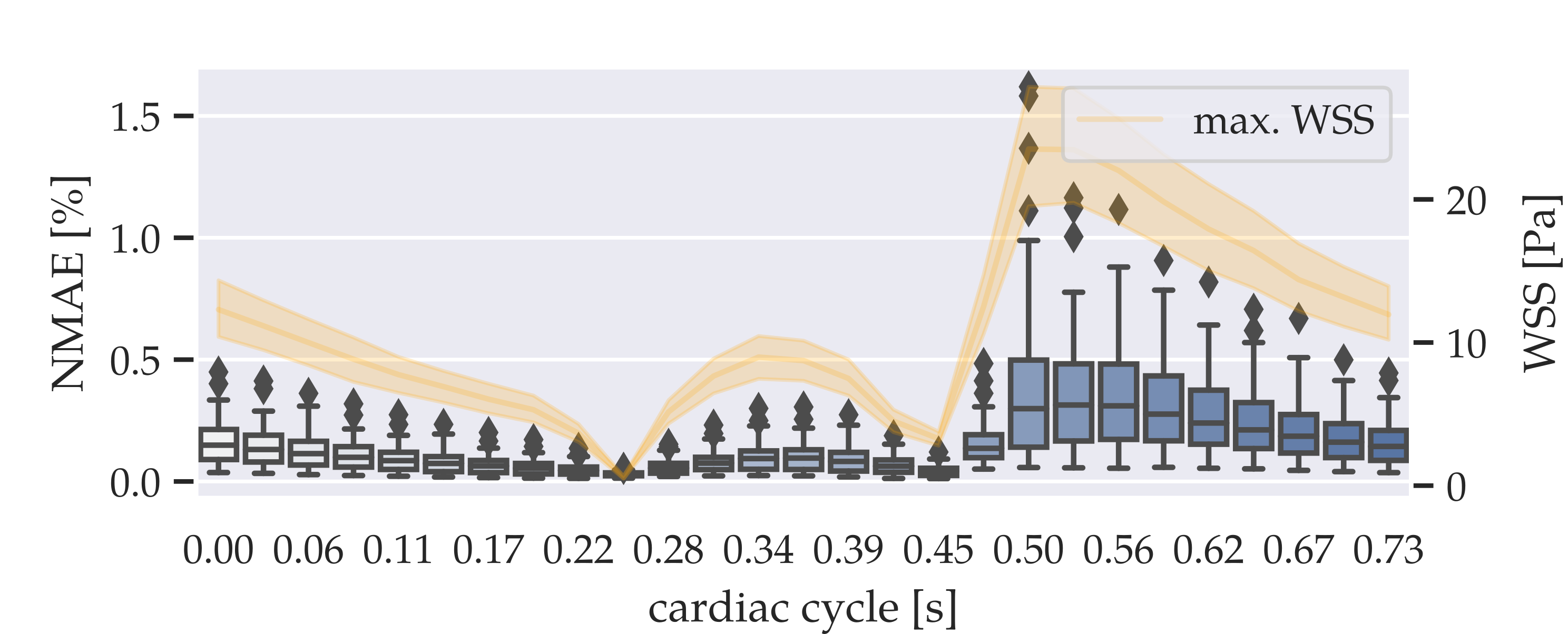}
    \caption{\textbf{Pulsatile} single-artery WSS prediction error across the test split over time. NMAE is normalised by the maximum WSS magnitude over all samples in the test set over time (indicated in yellow) which follows a pulsatile waveform.}
    \label{plt:temporal}
\end{figure}

%% file: plots/extrapolation.tex
\begin{figure}
    \centering
    \includegraphics[width=\columnwidth]{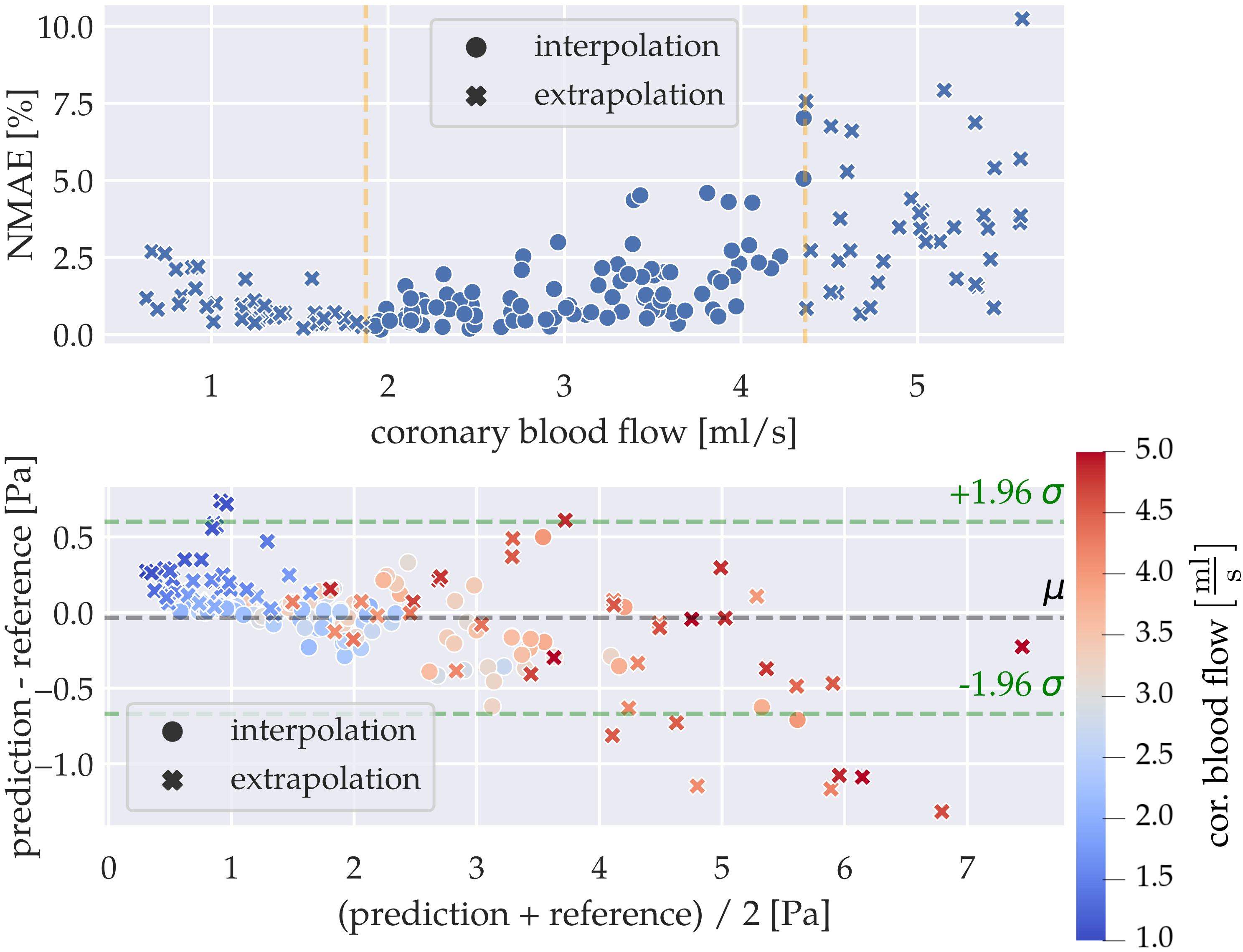}
    \caption{\textbf{Conditional, pulsatile} single-artery WSS prediction accuracy, subject to changing coronary blood flow boundary condition. Scatter plot (top) shows NMAE over the boundary condition value. Bland-Altman plot (bottom) shows the difference between neural-network prediction and ground-truth reference over their average, collapsed into a scalar value per artery by taking the mean over xyz-components, time, and mesh vertices. The mean of the difference is denoted by $\mu$ and the standard deviation by $\sigma$. GEM-GCN is trained on boundary conditions in {$[1.87, 4.36]$ $\frac{\text{ml}}{\text{s}}$}. Beyond, neural-network predictions are extrapolated.}
    \label{plt:extrapolation}
\end{figure}

%% file: figures/remeshing.tex
\begin{figure}
    \centering
    \includegraphics[width=\columnwidth]{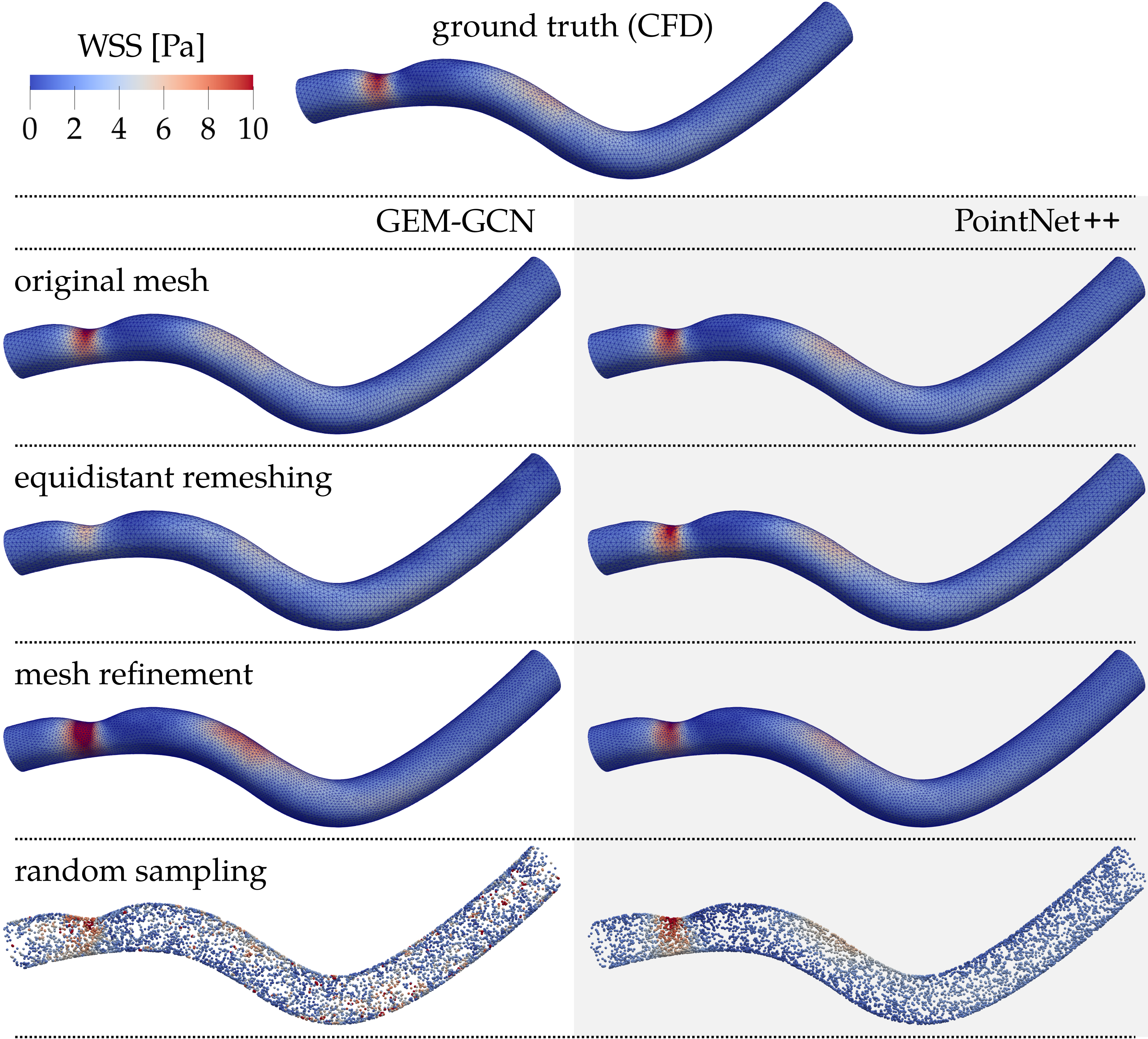}
    \caption{\textbf{Sensitivity to remeshing.} GEM-GCN (left column) and PointNet\text{++} (right column) trained on the original CFD mesh and evaluated on a differently remeshed artery wall $\partial\Omega$.}
    \label{fig:remeshing}
\end{figure}

%% file: figures/real.tex
\begin{figure}
    \centering
    \includegraphics[width=\columnwidth]{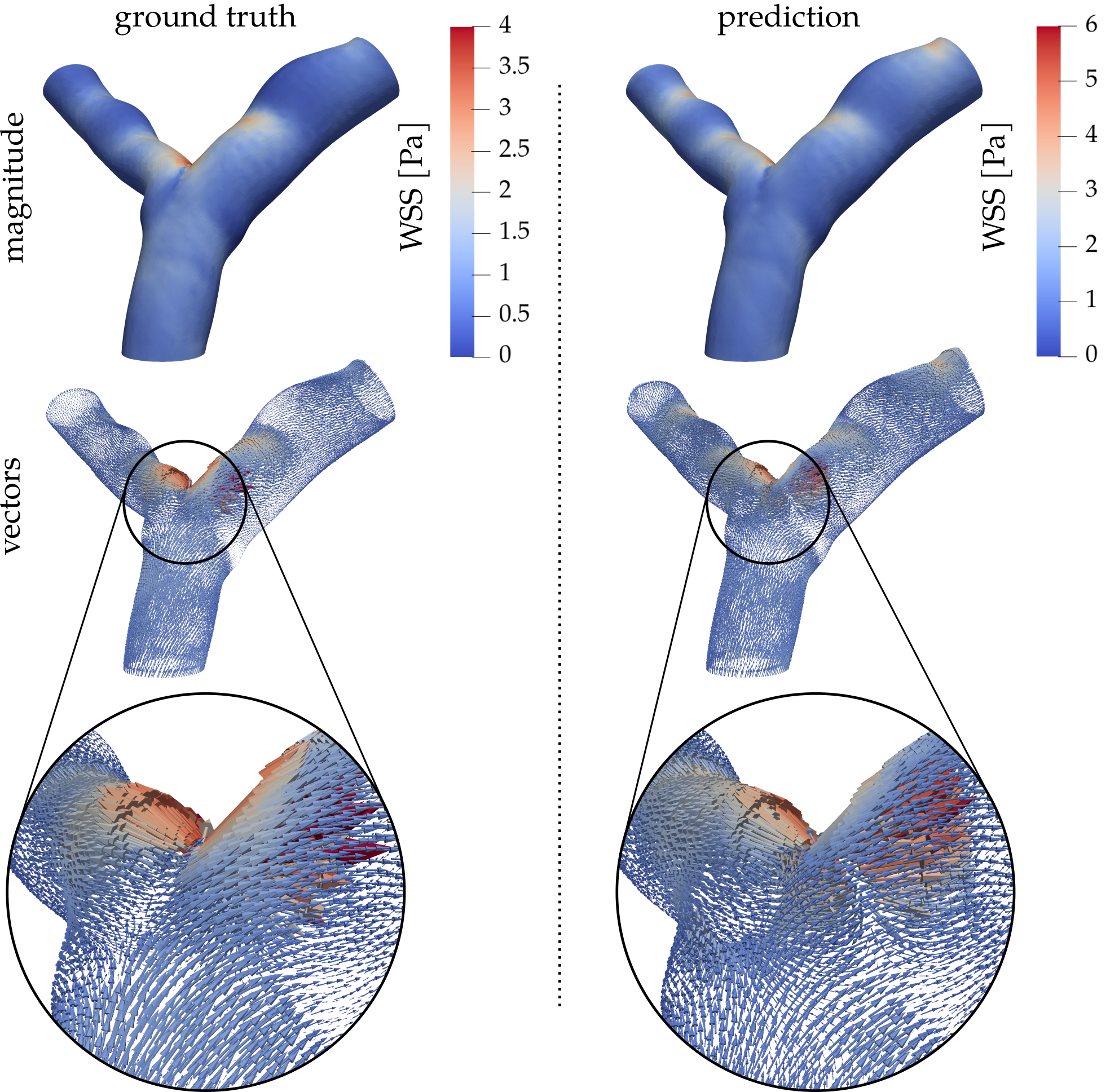}
    \caption{\textbf{WSS prediction for patient-specific} left main coronary bifurcation. Ground truth (left) versus GEM-GCN prediction (right). To produce these results, GEM-GCN is trained purely on the synthetic (steady-flow) bifurcating-artery dataset. Note that the colourbars are in different scales to facilitate qualitative comparison. The colour and size of the WSS vectors scale with magnitude.}
    \label{fig:real}
\end{figure}

%% file: content/conclusion.tex
\section{Discussion and conclusion}

We have presented an $\SE(3)$-equivariant GCN for the prediction of hemodynamic fields, operating on high-resolution surface mesh representations of the artery wall. Our results show that our method can learn to accurately predict vertex-wise, vector-valued, steady as well as pulsatile WSS in single and bifurcating, synthetic coronary arteries. Furthermore, results suggest that models can learn to inter- and extrapolate between and beyond coronary blood flow boundary conditions in the training set. Once trained, evaluation on unseen shapes takes less than 5 s, including geometric pre-processing. Our neural network is robust and flexible enough to be applied to a wide variety of different surface meshes. Fast estimation of physical quantities on meshes could benefit biomedical engineering applications where simulation data is abundant due to iterative fast prototyping, e.g. stent placement, but remains largely unused beyond a particular iteration. In this context, deep neural networks could estimate quantities of interest (e.g. oscillatory shear index (OSI)) for prototyping sub-iterations or supply initialisation for numerical solvers to speed up convergence of simulations. \revvision{In order for our method to be safely applied in clinical practice, it should be clinically validated and embedded in a pipeline with humans-in-the-loop, as is common practice~\cite{TaylorPetersen2023}. A core assumption of our method is that 3D patient models can be accurately extracted from medical images. Thus, the pipeline has to factor in segmentation variability caused by human or machine~\cite{KirisliSchaap2013,BergVoss2018} and resulting simulation variability~\cite{BergVoss2019} as well as simulation variability due to different CFD practitioners~\cite{ValenSendstadBergersen2018}. Modern segmentation methods can provide uncertainty estimates~\cite{ThibeauSutreAlblas2023}. In future work, it would be interesting to investigate if we can integrate this segmentation uncertainty in our method so as to compensate for variability in hemodynamics estimation.}

In contrast to previous works on hemodynamics estimation using deep learning, our method does not require projection to a 1D or 2D domain~\cite{ItuRapaka2016,SuZhang2020,GharleghiSamarasinghe2020,GharleghiSowmya2022,FerdianDubowitz2022}, does not disregard connectivity and curvature of the artery wall~\cite{LiangMao2020,LiWang2021}, and is independent of the embedding of the mesh in Euclidean space~\cite{MoralesFerezMill2021}. Instead, we operate natively on the geometric representation of the artery. We have demonstrated in Sec.~\ref{sec:method} how to exploit rotational and translational symmetry in our problem by an end-to-end $\SE(3)$-equivariant neural network.
In contrast, PointNet\texttt{++} (Sec.~\ref{subsec:baselines}) operates in 3D Euclidean coordinate space in which the geometric artery models are expressed. Thus, PointNet\texttt{++} is implicitly conditioned on the embedding of the input mesh. The only way to correct for this in non-equivariant neural networks is to perform data augmentation during training, effectively adding redundancy. We have demonstrated in Sec.~\ref{subsec:equivariance} that recovering the same accuracy as on registered input meshes requires longer training times and leads to lower accuracy. In fact, initial accuracy may never be fully recovered. Thus, when dealing with symmetric problems, GEM-GCN removes the need for roto-translational data augmentation and can lead to improved accuracy and data efficiency. Related to our approach are vector neurons~\cite{DengLitany2021}, a $\SE(3)$-equivariant point cloud network. Compared to our method, vector neurons is limited to a particular choice of $\SO(3)$-equivariant linear operation, while GEM-GCN uses an optimal gauge-equivariant linear operation. Finally, MeshCNN~\cite{HanockaHertz2019} has been heavily used for learning on meshes but defines its convolution to be invariant to rotation, sacrificing filter expressiveness compared to GEM-GCN. \revvision{Related works have modelled hemodynamics under consideration of the Navier-Stokes equations via physics-informed neural networks (PINN)~\cite{ArzaniWang2021,RaissiYazdani2020}. This line of works and ours represent two approaches with different use cases: iterative instance optimisation methods allow for incorporation of physics constraints but are slow, while generalising feed-forward methods appear black-box but are fast.}

Data-driven estimation of hemodynamic fields on the artery wall \revvision{can be data-hungry}~\cite{ArzaniWang2022}. To learn how geometry and hemodynamic fields relate, the neural network needs access to a sufficiently large and representative dataset\revvision{, especially when factoring in patient-specific boundary conditions}. In Sec.~\ref{subsec:steady}, we have quantified this \revvision{data} requirement for GEM-GCN. \revvvision{While for large, superficial arteries, a personalised waveform can be obtained via phase-contrast MRI or Doppler ultrasound, in many practical scenarios, e.g. for smaller, deeper arteries, a personalised waveform is difficult to obtain in a non-invasive manner. To account for the latter, i}\revvision{n this study, we have reduced the degrees of freedom of our cardiovascular boundary conditions to a single value which we use to scale a template waveform. \revvvisionout{This was based on the assumption that in many practical scenarios, e.g. arteries that cannot be easily imaged with ultrasound, a personalised waveform is difficult to obtain in a non-invasive manner.} In theory, we could increase the dimensionality of the boundary condition if we have enough training data, e.g. by parametrising patient-specific waveforms by a polynomial or spline representation. Furthermore, if patient-specific\revvvision{, measured} boundary conditions\revvvision{, such as waveforms and blood pressure,} are available, our neural networks can be trained with a parametrisation of these. Finding the optimal balance between complexity of the boundary condition and generalisation capabilities of the neural network is specific to the application and data availability. Studying this trade-off is an interesting direction for future research.} Neural networks have previously been found to do well at interpolating, but poorly at extrapolating training data~\cite{ArzaniWang2022}. However, we have demonstrated in Sec.~\ref{subsec:bct} that our method can to some extent extrapolate to different coronary blood flow boundary conditions. Our quantitative results have all been obtained on synthetic artery shapes and we have only provided preliminary results on a patient-specific artery in this work. Nevertheless, we have found that our method mildly generalises to real-life patient data. In future work, we aim to perform further validation on patient data with neural networks trained on synthetic data, which we can easily synthesise.

Additionally, we have investigated an important limitation of our method: accurate predictions require similar mesh connectivity, i.e. our method is sensitive to remeshing of the input surface. We hypothesise that this limitation can be alleviated by data augmentation. We find that PointNet\texttt{++} is more robust to remeshing, so it can be an option if heterogeneous mesh size is more important than $\SE(3)$ symmetry. Furthermore, we see this as an opportunity for discretisation-independent neural networks, e.g.~\cite{SharpAttaiki2022}.

Our method is based on the observation that WSS and pressure, in the laminar regime, depend in good approximation on artery wall shape and boundary conditions \textit{only}. This imposes a limitation on our work: in the turbulent regime, this hypothesis may be violated and thus our method would not be applicable. Furthermore, as in recent work by Gharleghi et al.~\cite{GharleghiSowmya2022}, we let our neural network output hemodynamic fields over a complete cardiac cycle discretised into fixed time steps simultaneously rather than iterating from one time step to the next, since the cardiac cycle is periodic and clinically relevant in its entirety. This is limiting if we want temporally finer resolved WSS estimation. Extending our approach to volumetric meshes and time-step simulation in future works could enable us to incorporate physical relations based on fluid velocity as additional inductive bias.

Even though we have collected a large dataset of hemodynamic simulations in arteries, we had to be selective with the types of simulations to run. We did not include pulsatile-flow fixed-inflow simulations for the bifurcating arteries, due to their extensive computational demand. In future work we could add them, but for now we already have pulsatile-flow varying-inflow simulations for the bifurcating arteries and fixed-inflow simulations would have limited additional value. \revvision{In our simulations, we only varied boundary conditions by average coronary blood flow and kept all else equal to be able to feasibly create a sufficiently large dataset. However, by design our method is not restricted to this simplified, parametrised boundary condition but can be conditioned on an arbitrary parametrisation. With access to a larger, more diverse dataset, we expect our method to be able to adapt to more complex boundary conditions, which is an interesting avenue for future research. In our simulations, we made several assumptions affecting the computed hemodynamics. Since our method mimics the relationship between input geometry and ground truth, as long as the data is consistent, we hypothesise that our method could be retrained to rapidly mimic the results of CFD simulations done by other practitioners. In future work, it would be good to investigate if the complexity of the CFD simulation affects our method's estimation performance positively or negatively, especially w.r.t. very detailed meshes.} \revvvision{It should be stated that there has been debate about the real-world, clinical utility of CFD for hemodynamics estimation~\cite{Kallmes2012,XiangTutino2014,CebralMeng2012,StrotherJiang2012,RobertsonWatton2012}. In practice, if in-vivo measurements of the desired ground truth are available, e.g. computed from 4D flow MRI, they could be used to train our neural network instead of simulated data. We plan to explore this approach in future work.}

In conclusion, we have shown that our proposed method can be a feasible plugin replacement for CFD for the task of fast, personalised estimation of hemodynamic quantities in high resolution on the artery wall.

%% file: appendix.tex
\input{appendix/proof}


\input{appendix/bifurcating}



%% file: appendix/proof.tex
\section{Proof of $\SE(3)$ equivariance (Prop.~\ref{prop:equivariance})}\label{app:proof}

An $\SO(3)$ representation $(\R^c, \rho)$ is a vector space $\R^{c}$ with an $\SO(3)$ action $\rho: \SO(3) \to \R^{c}\times \R^{c}$.  Let $\SO(2) \subset \SO(3)$ be the subgroup that leaves the z-axis invariant. The function $\rho:\SO(3)\to \R^{c}\times \R^{c}$ can be restricted $\res{\rho}{\SO(2)} : \SO(2) \to \R^c \times \R^c$ to give a representation of $\SO(2)$.

\begin{proposition}
    Choose input and output $\SO(3)$ features $(\R^{c_{\rm in}}, \rho^{\rm in})$ and $(\R^{c_{\rm out}}, \rho^{\rm out})$, which are also $\SE(3)$ representations that are invariant to translations. Choose a neural network consisting of GEM convolution, the pooling defined in Sec.~\ref{subsec:pooling}, gauge-equivariant activation functions~\cite{HaanWeiler2021}, and parameters such that the input and output $\SO(2)$ features are $(\R^{c_{\rm in}}, \res{\rho^\text{in}}{\SO(2)})$ and $(\R^{c_{\rm out}}, \res{\rho^\text{out}}{\SO(2)})$. For a mesh $\M = (\V, \F)$ including a choice of gauge, let $F_\M: \X(\V, \R^{c_{\rm in}}) \to \X(\V, \R^{c_{\rm out}})$ denote the neural network.
    
    For a transformation $g \in \SE(3)$, denote by $g\M$ the mesh where all the vertex positions are moved by the translation and rotation of $g$, and the normals and gauges are rotated by the rotation of $g$.
    
    For vertex $p \in \V$, let $w_{\M, p}\in \SO(3)$ be the rotation that maps the z-axis of (the ambient space) $\R^3$ to the normal vector of vertex $p$ and maps the x and y axes of $\R^3$ to the x and y axes on the tangent plane of vertex $p$, expressed in the choice of gauge. This is a basis transformation that maps from the global basis to a local basis at point $p$, consistent with the choice of gauge on the tangent plane. Applying this transformation for all vertices in $\SO(3)$ representation $(\rho, \R^c)$ gives an orthogonal linear transformation $\rho(w_\M): \X(\V, \R^c) \to \X(\V, \R^c)$. Now define the composition $\tilde F_\M = \rho^{\rm out}(w_\M)^{-1} \circ F_\M \circ \rho^{\rm in}(w_\M): \X(\V, \R^{c_{\rm in}}) \to \X(\V, \R^{c_{\rm out}})$. This composition is equivariant:
    \begin{equation}
        \rho^{\rm out}(g) \circ \tilde F_\M = \tilde F_{g\M}\circ  \rho^{\rm in}(g) \quad \forall g \in SE(3)
    \end{equation}
\end{proposition}
\begin{proof}
    The network only depends on the mesh through the intrinsic quantities of the parallel transport and the logarithmic map, which are equal in $g\M$ and $\M$ expressed in the respective gauges. In particular, $g$ preserves distances and angles, so the neighbourhoods $N(p)$ remain fixed under $g$. Thus, the network is invariant $F_\M=F_{g\M}$. Furthermore, as the gauge rotates with the transformation, if $r$ is the rotational part of $g$, then $w_{g\M, p}=w_{\M, p}r^{-1}$ and thus $\rho(w_{g\M})=\rho(w_\M)\circ \rho(g^{-1})$. Filling this in leads to
    \[
    \tilde F_{g\M}=\rho^{\rm out}(g) \circ \rho^{\rm out}(w_\M)^{-1} \circ F_\M \circ \rho^{\rm in}(w_\M)\circ \rho^{\rm in}(g^{-1})  
    \]
\end{proof}
\begin{remark}
    In the above, we chose the gauge of the transformed mesh $g\M$ to equal the rotated gauge of the original mesh $\M$. By construction, GEM-GCN is equivariant to the choice of gauge, so any argument that holds for this case extends to the general case as well.
\end{remark}

\begin{corollary}
    GEM-GCN together with input features defined in Sec.~\ref{sec:method} is $\SE(3)$-equivariant.
\end{corollary}
\begin{proof}
    The input features defined in Sec.~\ref{subsec:features} can be expressed vertex-wise as a $3 \cdot 3 \cdot 3$ dimensional $\SO(3)$ representation, given by the elements of three ($3 \times 3$) matrices:
    \begin{align*}
        m_\M^1(p) &= \sum\limits_{\B_r(p) \cap \V} \vec{v}_{p \to q} \vec{v}_{p \to q}^\T \\
        m_\M^2(p) &= \sum\limits_{\B_r(p) \cap \V} \vec{n}_q \vec{n}_q^\T \\
        m_\M^3(p) &= \sum\limits_{\B_r(p) \cap \V} \vec{v}_{p \to q} \vec{n}_q^\T
    \end{align*}
    where $\vec{v}_{p \to q} \in \R^3$ is the vector pointing from $p$ to $q$ and $\vec{n}_q$ is the vertex normal at $q$. Combined, these form a feature $m_\M \in \X(\V, \R^{27})$ with a $\SO(3)$ representation that acts on each matrix by conjugation: $\rho(g)(m)=gmg^T$. This feature is equivariant: $m_{g\M}=\rho(g)m_\M$. When this feature is used as an input to the network, the output is equivariant by Prop.~\ref{prop:equivariance}:
    \[
        \tilde F_{g\M}(m_{g\M})=\rho^{\rm out}(g)(\tilde F_\M(m_\M))
    \]
\end{proof}

%% file: appendix/bifurcating.tex
\section{\revision{Bifurcating artery synthesis}}\label{app:bifurcating}

The artery centerline of the parent vessel, PMV followed by DMV, is developed along seven control points and branches off into the child vessel SB at the fourth control point. The control points are evenly distanced $4$~[mm] apart. We construct the bifurcation in the y-z plane of a generic 3D coordinate system and sample two angles from the atlas~\cite{MedranoGraciaOrmiston2016} which together fully describe the bifurcation:
\begin{itemize}
	\item $\beta \sim \N(\mu_\beta, \sigma_\beta^2)$ with mean $\mu_\beta = 78.9^\circ$ and standard deviation $\sigma_\beta = 23.1^\circ$ which is the angle between centerlines of the branches DMV and SB and
	\item $\beta' \sim \N(\mu_{\beta'}, \sigma_{\beta'}^2)$ with mean $\mu_{\beta'} = 61.5^\circ$ and standard deviation $\sigma_{\beta'} = 21.5^\circ$ which is the angle between the bisecting line of the bifurcation and the centerline of SB.
\end{itemize}
The angle $\beta'$ describes how much the bifurcation is skewed towards the child branch (Fig.~\ref{fig:datasets}). We place the control points so that the angle between the line connecting the fourth and fifth point and the z-axis is $\beta'$ for SB and $\beta - \beta'$ for DMV. For a more realistic curvature, the angles between the lines connecting the other control points and the z-axis are linearly inter- and extrapolated starting from zero at the origin. To add curvature in x-direction, we sample a third angle $\gamma$ from the atlas:
\begin{itemize}
	\item $\gamma \sim \N(\mu_\gamma, \sigma_\gamma^2)$ with mean $\mu_\gamma = 9.5^\circ$ and standard deviation $\sigma_\gamma = 21.5^\circ$ which is the angle at which the PMV centerline enters the bifurcation plane.  
\end{itemize}
We place the control points so that the angle between the line connecting the third and fourth point and the z-axis is $\gamma$ while linearly inter- and extrapolating the angles between the lines connecting the other control points and the z-axis, starting from zero. To avoid unrealistic curvature, none of these angles must exceed $90^\circ$. The same (constant) curvature extends to both DMV and SB. It is anatomically unlikely for the LCX to curve upwards, so we restrict the SB to curve downwards. To arrive at the final centerline, the branching centerline path is smoothed using non-uniform rational basis splines (NURBS).

We model the vessel lumen with ellipse contours that are arbitrarily oriented in the plane normal to the centerline-curve tangent. The lumen radii are drawn from the coronary atlas~\cite{MedranoGraciaOrmiston2016}:
\begin{itemize}
	\item $r_\text{PMV} \sim \N(\mu_{r_\text{PMV}}, \sigma_{r_\text{PMV}}^2)$ with mean $\mu_{r_\text{PMV}} = 1.75$~[mm] and standard deviation $\sigma_{r_\text{PMV}} = 0.4$~[mm]
	\item $r_\text{DMV} \sim \N(\mu_{r_\text{DMV}}, \sigma_{r_\text{DMV}}^2)$ with mean $\mu_{r_\text{DMV}} = 1.6$~[mm] and standard deviation $\sigma_{r_\text{DMV}} = 0.35$~[mm]
	\item $r_\text{SB} \sim \N(\mu_{r_\text{SB}}, \sigma_{r_\text{SB}}^2)$ with mean $\mu_{r_\text{SB}} = 1.5$~[mm] and standard deviation $\sigma_{r_\text{SB}} = 0.35$~[mm]
\end{itemize}
Medrano-Gracia et al. empirically show that the measured lumen diameters coincide best, i.e. at the lowest root mean square error $\varepsilon$ across samples, with a bifurcation law of the form
\begin{equation*}
	(d_\text{PMV})^a = (d_\text{DMV})^a + (d_\text{SB})^a + \varepsilon
\end{equation*}
where $a = 2.4$~\cite{MedranoGraciaOrmiston2017}. As threshold we use the empirical root mean square error $\varepsilon = 0.165$ for the Huo-Kassab
bifurcation law $a = \frac{7}{3}$, since it is the bifurcation law with the closest value $a$ reported in \cite{MedranoGraciaOrmiston2017}. Accordingly, we choose values so that $\varepsilon \leq 0.165$ with the constraints that
\begin{itemize}
	\item $r_\text{PMV} < r_\text{DMV}$ or $r_\text{DMV} < r_\text{SB}$, based on the intuitions that the parent vessel should be larger than the child vessel and should not grow after a bifurcation and  
	\item $\frac{r_\text{SB}}{r_\text{DMV}} < 0.4$ according to empirical evidence from the atlas.
\end{itemize}
We observe
that vessel diameter decreases approximately linearly with vessel length in the relevant interval and linearly decrease it towards the end to 87.5~\% its initial size. To give the lumen a more realistic, non-smooth texture, we draw the contour ellipses' semi-minor and semi-major axes from a uniform noise distribution $\U(r - \delta, r + \delta)$ where $\delta = r \eta$ and $\eta = 5$~\%.


\input{tables/datasets}

%% file: tables/datasets.tex
\begin{table}[h]
	\begin{center}
    	\caption{\textbf{Dataset overview.} We run CFD simulations for synthetic single and bifurcating arteries for steady flow with fixed boundary condition, pulsatile flow with fixed boundary condition, and pulsatile flow with variable boundary conditions.}
    	\renewcommand{\arraystretch}{1.3}
    	\begin{tabular}{@{}lcc@{}}
    		\toprule
    		& Single arteries & Bifurcating arteries\\
    		\midrule
    		Steady flow & \checkmark & \checkmark\\
    		Pulsatile & \checkmark & \\
    		Conditional & \checkmark & \checkmark\\
    		\bottomrule
    	\end{tabular}
    	\label{tab:datasets}
	\end{center}
\end{table}